\documentclass[lettersize,journal]{IEEEtran}
\usepackage{amsmath,amsfonts}
\usepackage{algorithmic}
\usepackage{algorithm}
\usepackage{array}
\usepackage[caption=false,font=normalsize,labelfont=sf,textfont=sf]{subfig}
\usepackage{textcomp}
\usepackage{stfloats}
\usepackage{url}
\usepackage{verbatim}
\usepackage{graphicx}
\usepackage{hyperref}
\usepackage{cite}
\usepackage{booktabs}
\usepackage{multirow}
\usepackage{caption}
\usepackage{float}
\usepackage{bbding}

\usepackage{xcolor} 

\hypersetup{
    colorlinks=true,    
    linkcolor=blue,     
    filecolor=blue,     
    urlcolor=blue,      
    citecolor=blue,     
}

\begin{document}

\title{SA-Homo: Scale Adaptive Homography Estimation for Scale Variation Scenarios}

\author{Shangxuan Xie,
        Haifeng Wu,
        Yuhang Wang,
        Huarong Jia,
       Wen Li \textsuperscript{*}

    \IEEEcompsocitemizethanks{
    \IEEEcompsocthanksitem Shangxuan Xie, Haifeng Wu, and Wen Li (Corresponding Author) are with University of Electronic Science and Technology of China (UESTC), Chengdu, China (email: shangxuanx330@gmail.com; haifengwu205@gmail.com; liwenbnu@gmail.com.)
    \IEEEcompsocthanksitem Yuhang Wang is with University of Chinese Academy of Sciences, Beijing, China (email: yhw\_cx@163.com).
    \IEEEcompsocthanksitem Huarong Jia is with Beijing Institute of
    Technology, Beijing, China (email: 15810687706@163.com).
    }}

\markboth{Journal of \LaTeX\ Class Files,~Vol.~14, No.~8, August~2021}%
{Shell \MakeLowercase{\textit{et al.}}: A Sample Article Using IEEEtran.cls for IEEE Journals}

\IEEEpubid{0000--0000/00\$00.00~\copyright~2021 IEEE}
\maketitle

\begin{abstract}
Homography estimation, as one of the fundamental problems in computer vision, remains challenged by scale variation scenarios where image pairs potentially exhibit significant scale discrepancies. Existing deep learning frameworks frequently suffer from a significant performance degradation in such cases, as they rely on limited displacement assumptions and local feature consistency that might not hold under large scale gaps. In this paper, we propose SA-Homo, a novel scale-adaptive homography estimation framework designed to achieve robust alignment across a wide range of scale discrepancy ratios. We adopt a hierarchical scale alignment strategy that transitions from the global perspective with a heavy module to a local perspective with a light module. Specifically, we introduce the Scale-aware Discrepancy Bridging Module (SDBM) for initial alignment, which utilizes a Multi-scale Linear Attention Cascade (MLAC) to capture long-range dependencies and mitigate feature inconsistencies, along with a global Cross-scale Similarity Matrix Block (CSMB) for scale robust correlation representation. Once the initial scale gap is bridged, a lightweight Iterative Homography Estimation Refinement Module (IHERM) progressively polishes the result using local correlations. To facilitate this research, we contribute the HMSA dataset, a high-resolution, multi-modal satellite benchmark specifically tailored for scale-variant challenges. Extensive experiments demonstrate that SA-Homo maintains high precision even under 8$\times$ scale discrepancies, outperforming state-of-the-art methods in both conventional scale-similar scenarios and challenging scale variation scenarios. Code and collected datasets are available at 
\href{https://github.com/shangxuanx330/SA_Homo}{https://github.com/shangxuanx330/SA\_Homo}.

\end{abstract}


\section{Introduction}

\begin{figure}[t]
    \centering
    \includegraphics[width=1\columnwidth]{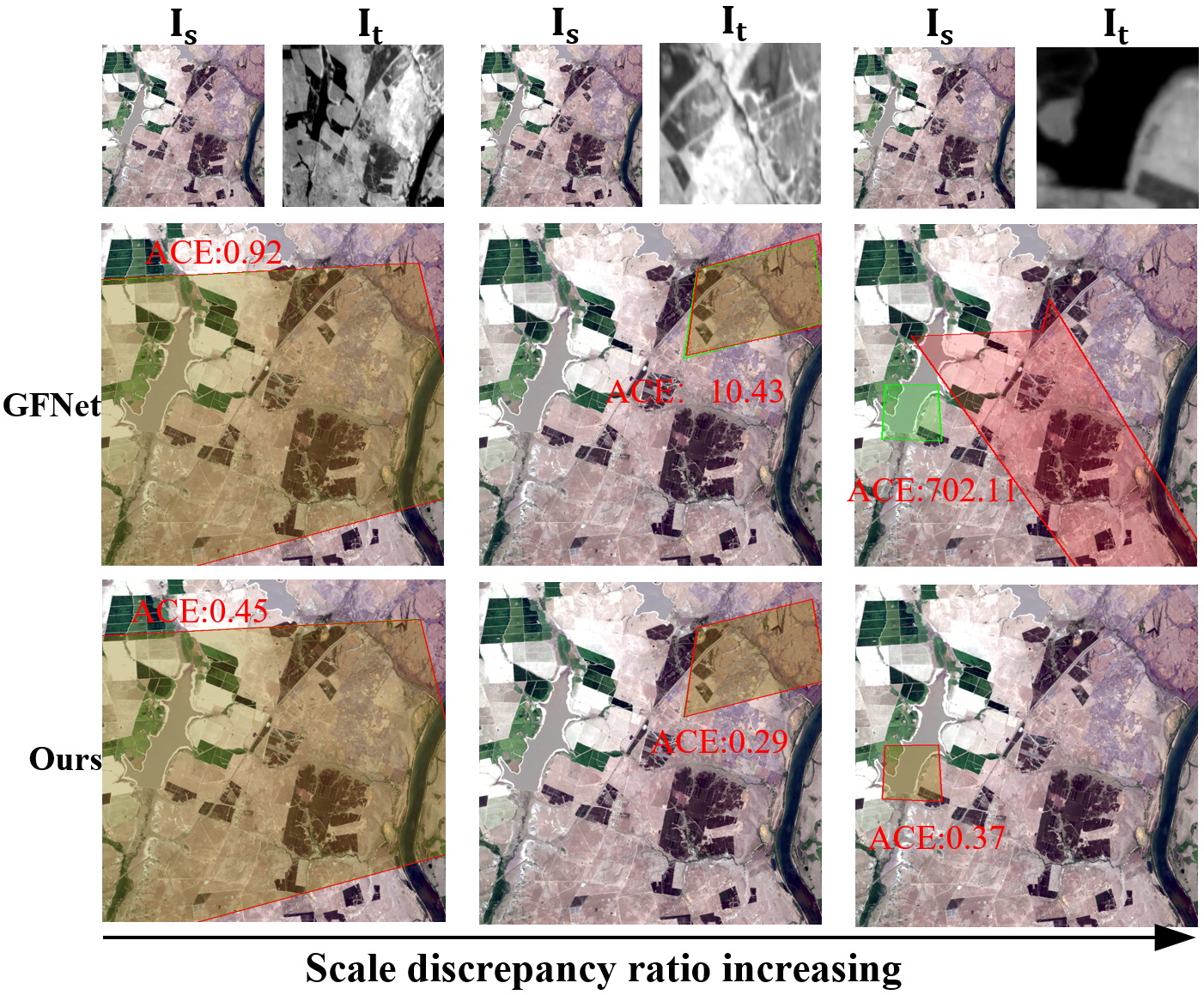}
    \caption{Homography estimation results under scale variation scenarios, where the scale discrepancy between the search image $I_s$ and template image $I_t$ varies (increasing from left to right). The red and green bounding boxes denote the prediction and ground truth, respectively. }
    \label{fig:intro}
\end{figure}

\IEEEPARstart{H}omography estimation is a fundamental task in computer vision that recovers the global projective transformation between two images taken from different viewpoints. It is widely applied in various multimedia processing and understanding tasks, such as image stitching~\cite{zhao2020fast,li2024seam,zarei2022megastitch}, multi-modal image fusion~\cite{li2023multimodal, zhou2024uncertainty, xie2024rcvs}, video stabilization~\cite{wang2025integrated,bradley2021cinematic,xu2022dut}, planar object tracking~\cite{zhan2022homography,xue2023smalltrack,linger2014aerial}. 

\begin{figure*}[t]
    \centering
    \includegraphics[width=1\textwidth]{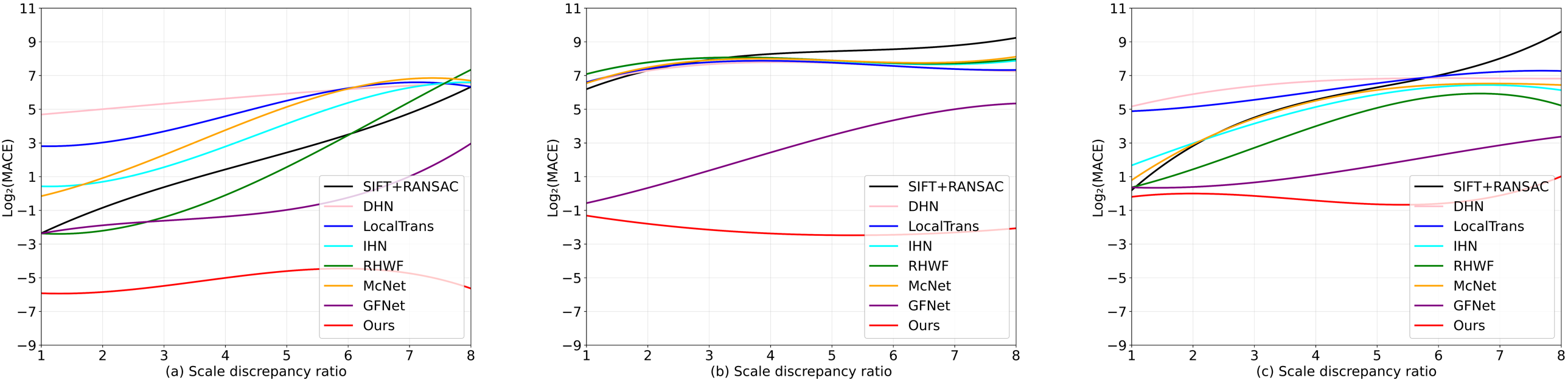}
    \caption{Impact of increasing scale discrepancy ratio on estimation accuracy across three datasets: (a) MSCOCO~\cite{lin2014microsoft}, (b) HMSA, and (c) RGB-NIR~\cite{brown2011multi}. Accuracy is quantified by \(\log_2(\text{MACE})\), where lower values indicate better performance. Notably, existing methods suffer from significant performance degradation as the scale discrepancy ratio increases, while our proposed method demonstrates robust performance under scale variation scenario.
     }
    \label{fig:mace_vs_sdr}
\end{figure*}

Recent years have witnessed significant progress in homography estimation, particularly with the emergence of deep learning-based methods. Early deep homography estimation methods directly regress homography parameters from stacked image pairs by designing various network architectures~\cite{detone2016deep,erlik2017homography,zhou2019stn}. Later works further improve the estimation accuracy by adopting iterative alignment paradigms, in which the predictions from the previous iteration are used to progressively warp the original images or feature maps, thereby gradually refining the alignment results~\cite{chang2017clkn,zhao2021deep,zhao2021image,cao2022iterative,cao2023recurrent,zhu2024mcnet,zhang2025adapting,le2020deep,shao2021localtrans}. However, existing works primarily consider image pairs with limited displacement and similar scales. When confronted with input pairs exhibiting varying scale discrepancies, state-of-the-art methods exhibit clear performance degradation (as shown in Fig.~\ref{fig:intro}). As illustrated in Fig.~\ref{fig:mace_vs_sdr}, previous methods suffer from significant performance degradation in scale variation scenarios. They either struggle to converge or exhibit a drastic performance decline as the scale discrepancy increases.

\IEEEpubidadjcol
The performance degradation of existing homography estimation methods in scale variation scenarios can be mainly attributed to two factors. First, image pairs in scale variation scenarios often involve considerable scale discrepancies, leading to large displacements and inconsistent local appearances between corresponding points. These effects violate the limited-displacement assumption adopted by many existing methods, thereby weakening their robustness under significant scale changes~\cite{chang2017clkn,zhao2021deep,zhu2024mcnet}. Second, the challenging nature of scale variation scenarios makes networks more prone to producing outliers during correspondence establishment. Since most existing methods do not explicitly suppress the influence of such outliers during iterative optimization, their estimation accuracy can be further compromised~\cite{zhang2025adapting}. These observations highlight the need for a homography estimation framework that can explicitly cope with large displacements and inconsistent local features, while effectively reducing the adverse impact of outliers.

To address the aforementioned issues, we propose a novel scale-adaptive homography estimation framework designed for scale variation scenarios, termed SA-Homo which based on hierarchical scale alignment strategy. Our approach begins with a Scale-aware Discrepancy Bridging Module (SDBM) for robust initial scale alignment. Within SDBM, we design a Cross-scale Similarity Matrix Block (CSMB) to compute and represent correlations tailored to scale variation scenarios. CSMB constructs global-range correlation representations to handle large displacements and builds cross-scale correlations to alleviate scale-induced local appearance inconsistencies. In addition, to further mitigate local feature inconsistency and obtain more scale-robust features, the SDBM incorporates multi-scale linear attention cascades to fuse information from multiple receptive fields. Finally, RANSAC~\cite{fischler1981random} is adopted to filter out unreliable correspondences for robustness. Although SDBM provides robust initial alignment, it operates on highly downsampled feature maps (\emph{e.g.}, $16\times$ downsampling) to alleviate the computational burden, which limits its ability to achieve sub-pixel-level accuracy. Consequently, we utilize a lightweight Iterative Homography Estimation Refinement Module (IHERM) to enable fine-grained iterative estimation. With the scale discrepancy significantly reduced by the SDBM, it is not necessary for the IHERM to utilize a heavy module to perform global correspondence search. Instead, IHERM can be more lightweight and focus on processing correlations within a local range. The proposed hierarchical scale alignment strategy, which transitions from global to local perspectives and from heavy to lightweight modules, delivers scale robustness while preserving computational efficiency. To evaluate homography estimation in practical scenarios where scale variations frequently occur, we construct a dataset named HMSA (High-resolution, Multi-modal, Satellite, Aligned dataset) from publicly available Landsat8 satellite imagery\cite{roy2014landsat}. HMSA consists of 11,837 registered visible-infrared satellite image pairs at a resolution of 1152$\times$1152. Extensive experiments conducted on both our proposed dataset and public benchmarks demonstrate the effectiveness and robustness of our method in both challenging scale variation and conventional scale-similar scenarios. In summary, our main contributions are as follows:

\begin{itemize}
\item We introduce SA-Homo, a scale-adaptive homography estimation framework that achieves accurate estimation under scale discrepancies of up to $8\times$, while maintaining high computational efficiency through a hierarchical scale alignment strategy that progressively transitions from heavy to lightweight modules.

\item We develop a Scale-aware Discrepancy Bridging Module (SDBM) to provide robust initial alignment under scale variation scenarios. It incorporates a Cross-scale Similarity Matrix Block (CSMB) to establish global cross-scale correlations for handling large displacements and local feature inconsistency issues, and further employs outlier removal algorithm to improve robustness.

\item Extensive experiments across multiple datasets validate the effectiveness and robustness of our method across conventional scale-similar and challenging scale variation scenarios. We also contribute the HMSA dataset, which serves to facilitate the exploration of deep homography estimation under scale variation scenario.
\end{itemize}

The rest of this paper is organized as follows. Related works are introduced in Section~\ref{related work}. In Section~\ref{method}, we detail the framework of SA-Homo and our data generation pipeline. In Section~\ref{experiments}, we introduce the relevant datasets and experimental setups, and report comprehensive comparative results from our experiments. In Section~\ref{conclusion}, we conclude the paper.

\begin{figure*}[t]
    \centering
    \includegraphics[width=1\textwidth]{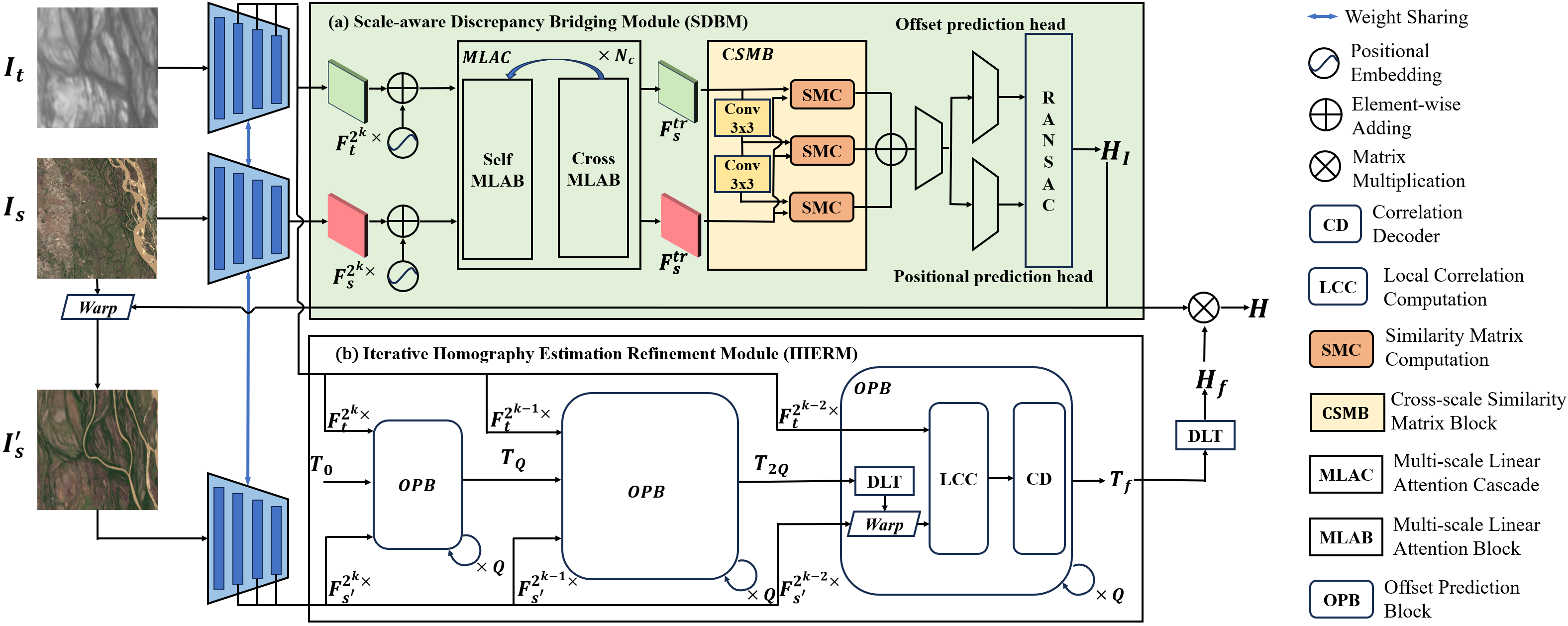}
       \caption{Overview of the proposed SA-Homo. The architecture consists of two main modules: (a) the Scale-aware Discrepancy Bridging Module (SDBM) and the (b) Iterative Homography Estimation Refinement Module (IHERM). The SDBM performs an initial alignment by capturing global correlations to significantly reduce scale discrepancy, thereby enabling the lightweight IHERM to progressively refine the estimation by focusing on local correlations. Specifically, multi-scale features \(\{F^{2^i \times}\}_{i=k, k-1, k-2}\) are extracted from both images using a shared feature extractor with downsampling factors of \(2^i\). The SDBM takes the features \(F_t^{2^k \times}\) and \(F_s^{2^k \times}\) as input to produce a initial homography matrix \(H_I\), which warps \(I_s\) into \(I_s^{\prime}\). Subsequently, multi-scale features are extracted from the warped image \(I_s^{\prime}\) and fed into the IHERM paired with the template features to iteratively estimate a refined homography \(H_f\). The final estimated homography is obtained by composing the two estimates: \(H = H_f \cdot H_I\).}
    \label{fig:Overall_framework}
\end{figure*}

\section{Related Work}
\label{related work}
\textbf{Traditional homography estimation.} Conventional homography estimation methods typically adopt a three-stage pipeline comprising feature detection, correspondence matching, and homography matrix computation. In the feature detection stage, representative hand-crafted approaches include the Scale-Invariant Feature Transform (SIFT)~\cite{lowe2004distinctive}, Speeded-Up Robust Features (SURF)~\cite{bay2006surf}, and Oriented FAST and Rotated BRIEF (ORB)~\cite{rublee2011orb}. More recently, learning-based feature detection methods have emerged, including SuperPoint~\cite{detone2018superpoint}, SOSNet~\cite{tian2019sosnet}, and D2-Net~\cite{dusmanu2019d2}. Furthermore, methods such as LoFTR~\cite{sun2021loftr} and EfficientLoFTR~\cite{wang2024efficient} have been proposed to perform feature detection and correspondence matching in a unified framework. Following the establishment of correspondences between image pairs, the homography matrix computation is conducted through a two-step procedure. First, outlier correspondences are filtered using robust estimation techniques such as RANSAC~\cite{fischler1981random}, IRLS~\cite{holland1977robust}, and MAGSAC~\cite{barath2019magsac}. Subsequently, the Direct Linear Transform (DLT)~\cite{hartley2003multiple} is applied to derive the final homography matrix. Traditional hand-crafted approaches offer the advantage of being training-free. However, the reliance on hand-crafted features introduces significant limitations, particularly under challenging conditions such as large viewpoint changes, cross-modal scenarios, and scale variation scenarios, which often result in substantial performance degradation.

\textbf{Deep homography estimation.} As a seminal work, DHN~\cite{detone2016deep} pioneered the field by utilizing a VGG-style framework to directly regress homographies from image pairs.  Subsequently, various approaches have been developed to handle different scenarios and improve estimation accuracy. For dynamic scene scenarios, \cite{zhou2019stn} incorporates a Spatial Transformer Network (STN) to align images, whereas MHN~\cite{le2020deep} introduces a learnable masking mechanism to mitigate interference from moving objects, and CodingHomo~\cite{liu2024codinghomo} leverages bitstream motion vectors as priors via mask-guided modules to suppress dynamic foregrounds for better performance. For cross-modal scenarios, DLKFM~\cite{zhao2021deep} projects images from distinct modalities into a shared latent space. For cross-resolution scenarios, LocalTrans~\cite{shao2021localtrans} leverages local transformers within a multi-scale network to capture correspondences, while CrossHomo~\cite{deng2024crosshomo} integrates super-resolution techniques to assist estimation and leverages a multi-modal channel shuffle strategy to better address cross-modality issues. Alongside scenario-specific designs, iterative refinement strategies have been widely adopted to further enhance estimation accuracy. Early approaches based on the Inverse Compositional Lucas-Kanade (IC-LK) algorithm~\cite{chang2017clkn,zhao2021deep} combine CNN-extracted features with pre-computed, non-learnable iterators. To overcome the limitation of non-learnable iterators, recent works have evolved towards fully learnable frameworks. IHN~\cite{cao2022iterative} proposes an end-to-end iterative architecture with a trainable iterator, effectively circumventing the limitations inherent in non-learnable schemes~\cite{chang2017clkn,zhao2021deep}. Similarly, RHWF~\cite{cao2023recurrent} introduces a FocusFormer and a homography-guided warping strategy to achieve better performance. More recently, McNet~\cite{zhu2024mcnet} employs a multi-scale correlation searching strategy to achieve superior performance. In a recent advancement, GFNet~\cite{zhang2025adapting} integrates a frozen DINOv2 encoder with a sparsified implementation of pixel-level dense matching~\cite{edstedt2024roma} to achieve high-precision estimation. GFNet initializes the optical flow field from the patch-level correlation volume through parameter-free soft-argmax operation, which may limit its effectiveness in handling complex scenarios with significant scale variations.

While previous methods have achieved substantial progress across various scenarios and demonstrated significant accuracy improvements, they are primarily designed for scale-similar image pairs and thus often encounter limitations under scale variation conditions. Unlike previous works, our proposed SA-Homo addresses homography estimation under more general and challenging scale variation scenarios.

\section{Method}
\label{method}
In scale variation scenarios, the scale discrepancy ratio between image pairs can span a wide range (e.g., from 1 to 8), potentially leading to substantial displacements and inconsistent local feature representations among corresponding points, rendering existing methods less effective. Within our hierarchical scale alignment strategy, these dual challenges are tackled by the Scale-aware Discrepancy Bridging Module  (SDBM). SDBM achieves initial homography estimation under scale variation, yielding $H_I$ for rapid scale alignment. With the scale discrepancy significantly reduced, an Iterative Homography Estimation Refinement Module (IHERM) focused on local correlation is used to iteratively align the features between $I_s^{\prime}$ and $I_t$, producing the cumulative refinement matrix $H_f$. The final estimation result of our model is computed as $H = H_f \cdot H_I$. The overall pipeline is illustrated in Fig.~\ref{fig:Overall_framework}, with detailed descriptions of the main modules provided in the subsequent sections.

\begin{figure}[t]
    \centering
    \includegraphics[width=1\columnwidth]
    {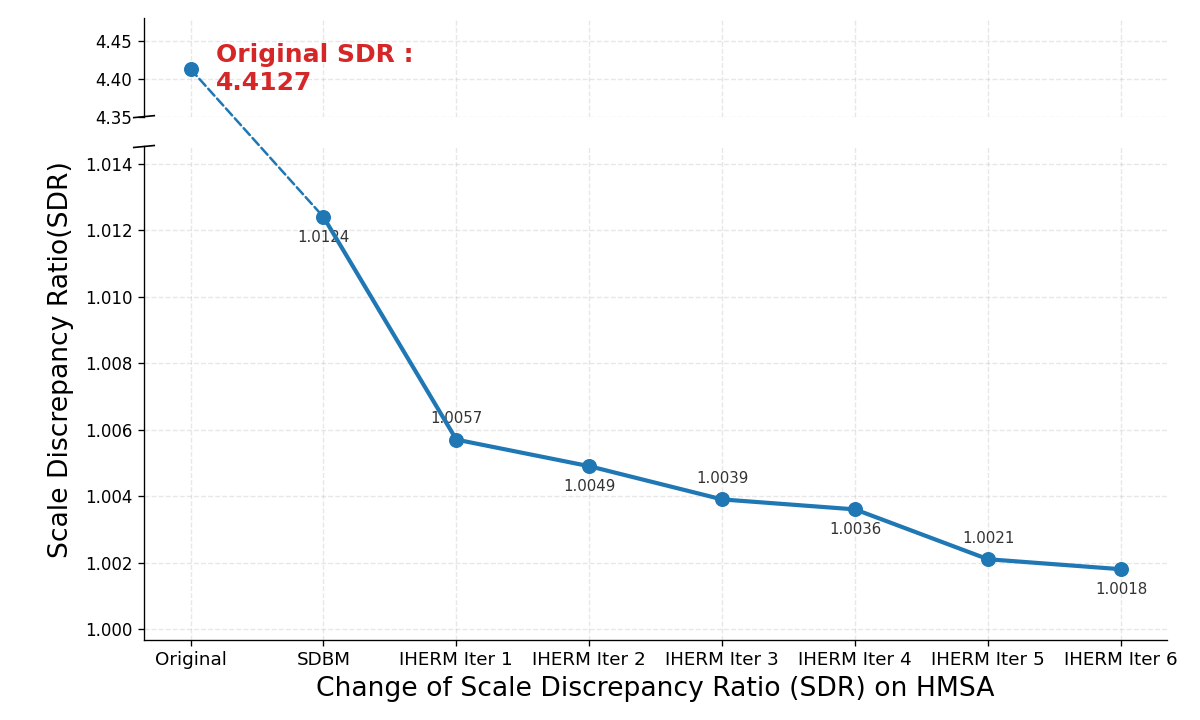}
    \caption{Visualization of the progressive evolution of the Scale Discrepancy Ratio (SDR) across our model. The SDBM significantly reduces the scale discrepancy (from 4.41 to 1.01 on average).}
    \label{fig:scale_changes_in_each_module_stage}
\end{figure}

\subsection{Scale-aware Discrepancy Bridging Module }
To enable robust initial alignment under scale variation scenarios, it is crucial to effectively address the large displacements and inconsistent local features across corresponding points caused by significant scale discrepancy. The former requires global scope correlation representation and prediction search range, while the latter benefits from endowing the model's features with more global context and the aggregation of cross-scale correlation information.

The framework of SDBM is shown in Fig.~\ref{fig:Overall_framework} (a), we first employ a Multi-scale Linear Attention Cascade (MLAC) to capture long-short range dependencies, facilitate interactions between feature maps, and aggregate multi-scale information. Building upon these enhanced features, we then introduce a Cross-scale Similarity Matrix Block (CSMB) to represent correlations, which are no longer constrained to local windows. Accordingly, SDBM explicit estimates the positions of $K$ grid-sampled keypoints based on similarity matrix output by CSMB. After obtaining the $K$ correspondences, we apply RANSAC \cite{fischler1981random} and DLT~\cite{hartley2003multiple} to filter out outliers and compute the initial homography $H_I$. $I_s$ is then warped by $H_I$ to get $I_s'$, performing the initial scale alignment.

\begin{figure}[t]
    \centering
    \includegraphics[width=0.9\columnwidth]
    {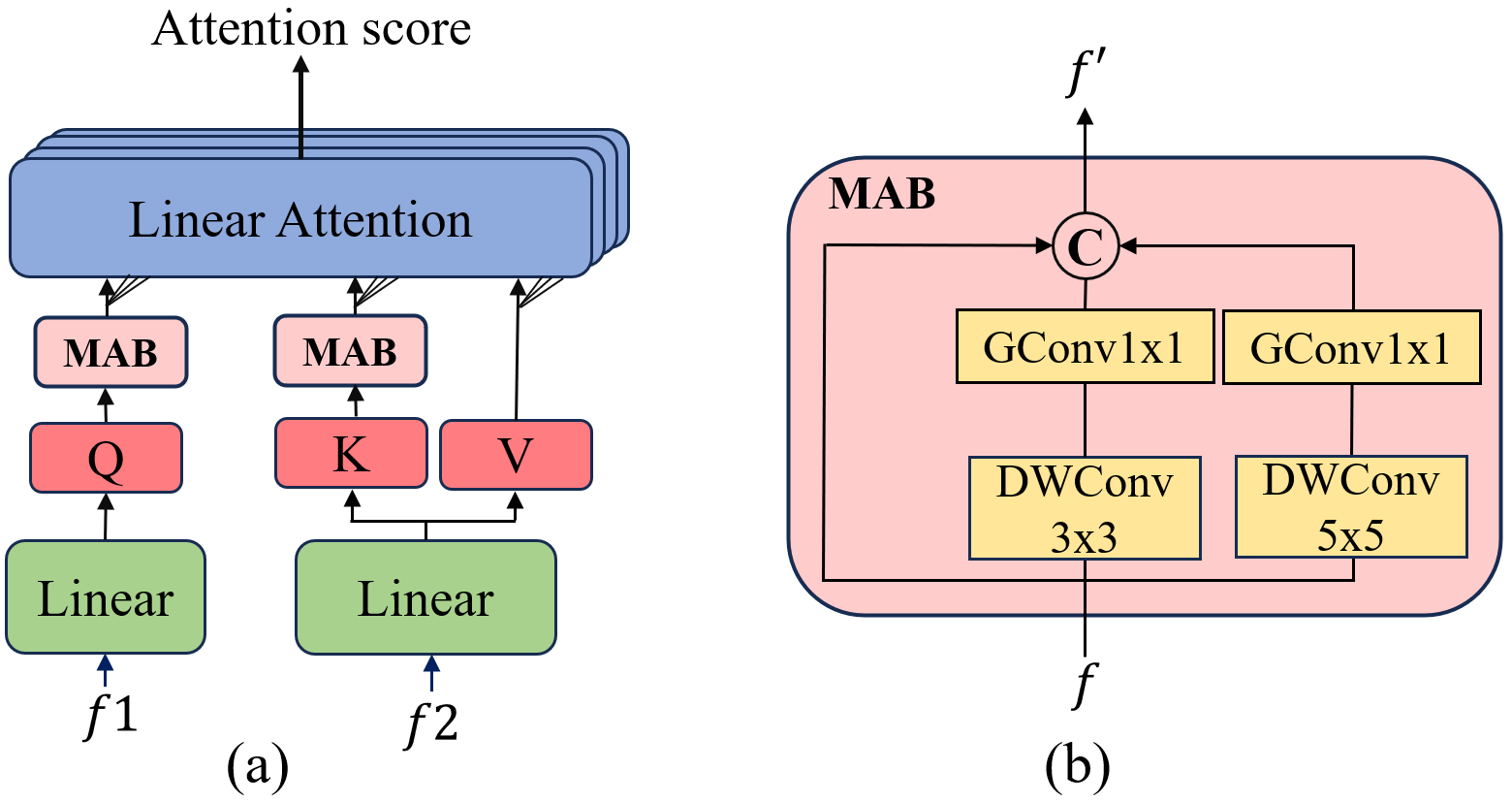}
    \caption{(a) Illustration of the Multi-scale Linear Attention mechanism. (b) Structure of the Multi-scale Aggregation Block (MAB).}
    \label{fig:MLA_AND_MAB}
\end{figure}

\textbf{Multi-scale Linear Attention Cascade (MLAC).}  We adopt MLAC~\cite{xie2025p2wnet} for feature enhancement, which has $N_c$ basic layers, and each layer consists of self and cross Multi-scale Attention Blocks (MLAB). The framework of MLAC is modified from the LOFTR module~\cite{sun2021loftr}, where the linear attention\cite{katharopoulos2020transformers} is substituted with multi-scale linear attention~\cite{cai2022efficientvit} to better capture multi-scale information for addressing scale variations. Multi-scale linear attention is an enhanced variant of linear attention that integrates a Multiscale Aggregation Block (MAB) before attention score computation to improve multi-scale capability. As illustrated in Fig.\ref{fig:MLA_AND_MAB} (b), the MAB utilizes depthwise convolutions \cite{howard2017mobilenets} combined with $1\times1$ group convolutions to extract different scale features with different kernel sizes, which are subsequently concatenated to form a multi-scale representation. The multi-scale linear attention score is then computed as follows:

\begin{equation} 
\text{Attention}(Q, K, V) = \frac{\phi(\varphi(Q)) \cdot (\phi(\varphi(K))^\top V)}{\phi(\varphi(Q)) \cdot \phi((\varphi(K))^\top \mathbf{1})},
\end{equation}
where $Q$, $K$, and $V$ denote query, key, and value, the $\phi(\cdot) = elu(\cdot)+1$, and the $\varphi(\cdot) = MAB(\cdot)$,

\textbf{Cross-Scale Similarity Matrix Block (CSMB).}
We propose CSMB to compute and represent the correlation between two images under scale variations. It is characterized by two key features: (i) To handle large displacement issue, we perform correlation computation in a global scope rather than restricting to local neighborhoods; (ii) To mitigate local feature inconsistencies arising from varying spatial correspondences, where a pixel in $I_s$ may correspond to a region of varying size in $I_t$ depending on the scale discrepancy ratio between the image pair, we explicitly evaluate cross-scale correlations by correlating $F_s^{tr}$ with multi-granularity features constructed from $F_t^{tr}$. Specifically, let $F_{t-1}^{tr}$, $F_{t-3}^{tr}$, and $F_{t-5}^{tr}$ represent the features obtained by applying zero, one, and two $3\times 3$ convolution blocks to the original $F_{t}^{tr}$, respectively. As illustrated in Fig.~\ref{fig:Overall_framework} (a), the cross-scale similarity matrix $S \in \mathbb{R}^{(h_s w_s)\times (h_t w_t)}$ is computed as:
\begin{equation}
S(i,j) = \sum_{k\in\{1,3,5\}}\frac{\langle F_s^{tr}(i),\, F_{t-k}^{tr}(j)\rangle}{D},
\end{equation}
where $\langle\cdot,\cdot\rangle$ denotes the inner product and $D$ is the feature dimension. $S(i,j)$ measures the similarity between the $i$-th token in $F_s^{tr}$ and the $j$-th token in $F_t^{tr}$. 

After obtaining $S$, we transform it into a soft assignment matrix to mitigate ambiguous many-to-one matches. Specifically, we employ the iterative Sinkhorn-Knopp algorithm~\cite{cuturi2013sinkhorn}, which alternately normalizes rows and columns to yield a assignment matrix. We initialize the process by scaling $S$ with a temperature parameter $\tau$:
\begin{equation}
A^{(0)} = S/\tau.
\end{equation}
Then, we implement the iterative scaling in the log-domain. For $t=0,\dots,T-1$, we alternately apply row-wise and column-wise log-normalization:

\begin{align}
A^{(t+\frac{1}{2})}_{ij} &= A^{(t)}_{ij} - \log\sum\nolimits_{j'}\exp\!\left(A^{(t)}_{ij'}\right), \\
A^{(t+1)}_{ij} &= A^{(t+\frac{1}{2})}_{ij} - \log\sum\nolimits_{i'}\exp\!\left(A^{(t+\frac{1}{2})}_{i'j}\right).
\end{align}
Finally, we obtain the soft assignment matrix by exponentiation:
\begin{equation}
\hat{S}=\exp\!\left(A^{(T)}\right),
\end{equation}
where $\hat{S}$ is sent to the position and offset prediction heads for keypoint prediction.

\textbf{Keypoint prediction and homography estimation.}
To predict the corresponding position in $I_s$ of the $K$ grid-sampled keypoints from $I_t$, we employ a prediction module consisting of position and offset prediction heads~\cite{yang2021siamcorners}. The position prediction head generates a heatmap $\mathbf{HM} \in \mathbb{R}^{K \times h_s \times w_s}$, while the offset prediction head produces an offset map $\mathbf{OM} \in \mathbb{R}^{2\times K \times h_s \times w_s}$. The predicted coordinates for the $i$-th keypoint are obtained as follows.

First, we extract coarse predictions $(X_i^c, Y_i^c)$   by identifying the position of maximum activation in the $\mathbf{HM}$ :
\begin{equation}
(X_i^{c}, Y_i^{c}) = \underset{(x,y)}{\text{argmax}}\mathbf{HM}(i, x, y),
\end{equation}
then, we obtain the refinement $(\Delta x_i, \Delta y_i)$ from $\mathbf{OM}$:
\begin{equation}
(\Delta x_i, \Delta y_i) = (\mathbf{OM}(0,i,X_i^c, Y_i^c), \mathbf{OM}(1,i,X_i^c, Y_i^c)),
\end{equation}
finally, the predicted coordinates $(X_i^{pred}, Y_i^{pred})$ for the $i$-th keypoint are computed as:
\begin{equation}
(X_i^{pred}, Y_i^{pred}) = F_n(X_i^{c}, Y_i^{c}) = 2^k \times (X_i^c + \Delta x_i, Y_i^c + \Delta y_i),
\label{eq:F_n}
\end{equation}
where $2^k$ represents the downsampling factor for SDBM.

\begin{figure}[t]
    \centering
    \includegraphics[width=1\columnwidth]
    {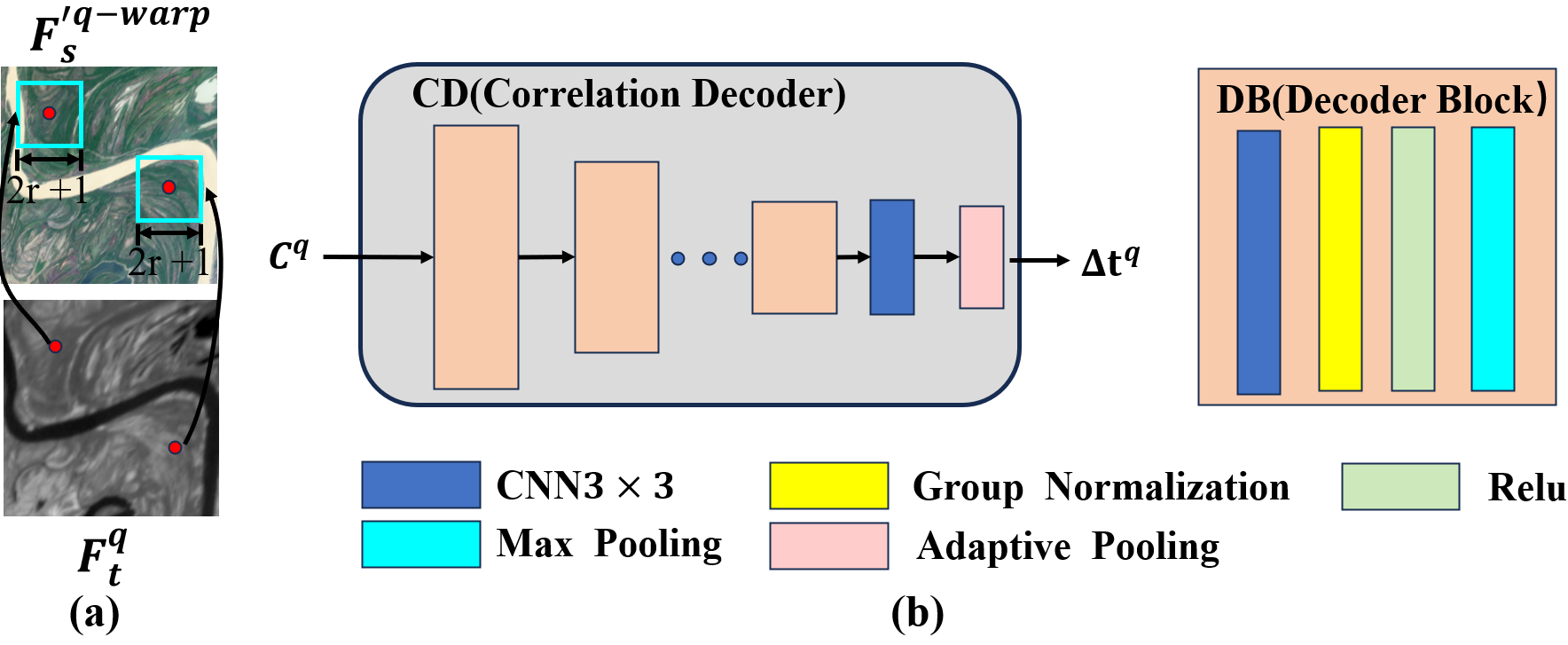}
    \caption{(a) Illustration of the local correlation computation. (b) The framework of Correlation Decoder.  }
    \label{fig:CCCD}
\end{figure}
\subsection{Iterative Homography Estimation Refinement Module}
To alleviate the computational burden, the SDBM module typically operates on highly downsampled feature maps, which limits its capability to achieve sub-pixel-level accuracy. With the scale discrepancy significantly reduced (as illustrated in Fig.~\ref{fig:scale_changes_in_each_module_stage}), computational components such as MLAC and CSMB are no longer required to address large scale discrepancy. The subsequent refinement can be efficiently processed by lightweight module architectures focusing on local correlation representations. To this end, we incorporate the Iterative Homography Estimation Refinement Module (IHERM), inspired by~\cite{zhu2024mcnet}, to further enhance model performance while preserving computational efficiency.

The overall architecture of IHERM is depicted in Fig.~\ref{fig:Overall_framework} (b). It accepts the multi-scale feature sets $\{F_t^{v \times}\}_{v \in \mathcal{V}}$ and $\{F_s^{\prime v \times}\}_{v\in \mathcal{V}}$ (where $\mathcal{V} = \{2^k, 2^{k-1}, 2^{k-2}\}$) as input, subsequently performing $Q$ iterative refinements at each scale to progressively estimation. In the IHERM framework, the homography matrix $H_i$ (where $i=1,...,3Q$) is parameterized by the offsets of four corner points, denoted as $T_i \in \mathbb{R}^{2\times2\times2}$ . Once $T_i$ is determined, the corresponding $H_i$ is solved via the Direct Linear Transform (DLT)~\cite{hartley2003multiple}. The iterative alignment process begins with an initialization of $T_0 = \mathbf{0}$. During the $q$-th iteration, as illustrated in Fig.~\ref{fig:Overall_framework}(b), the Offset Prediction Block (OPB) regresses a residual offset $\Delta t_q$. This residual is used to update the current estimate via $T_q = T_{q-1} + \Delta t_q$, which initializes the state for the $(q+1)$-th iteration. Below, we provide a detailed breakdown of the OPB's internal architecture, specifically focusing on the Correlation Computation and the Correlation Decoder.

\textbf{Local Correlation Computation.} 
For simplicity, we denote $F_t^q$ and $F_s^{\prime q}$ as the feature inputs to the OPB in the $q$-th iteration. Before computing the correlation, we utilize the current accumulated offset $T_{q-1}$ to obtain $H_q$ through DLT and warp $F_s^{\prime q}$ to obtain $F_s^{\prime q-warp}$. As illustrated in Fig.~\ref{fig:CCCD} (a), we compute the correlation volume $C\in \mathbb{R}^{(2r+1)\times (2r+1) \times h_s^q \times w_s^q}$ using the following formula:
\begin{equation}
C^q(u,v,x,y) = F_t^q(x,y)^T \cdot F_s^{\prime q\text{-warp}}(x+u, y+v),
\end{equation}
where $u,v \in \{-r, -r+1, \ldots, r-1, r\}$.

\textbf{Correlation Decoder.} The framework of Correlation Decoder (CD) is shown in Fig.~\ref{fig:CCCD} (b). It consists of several Decoder Blocks (DB), followed by a convolutional layer and an adaptive pooling layer. It takes reshaped correlation volume $C_{r}^{q} \in \mathbb{R}^{((2r+1)\times (2r+1)) \times h_s^q \times w_s^q}$ as input and predicts the offset residual $\Delta t_q \in \mathbb{R}^{2\times2 \times 2}$ in the $q$-th iteration.

\subsection{Loss Function} The overall loss function is formulated as:
\begin{equation}
\mathcal{L} = (\mathcal{L}_{\text{sim}} + \mathcal{L}_p + \alpha \mathcal{L}_o) + \mathcal{L}_I,
\label{eq:overall_loss}
\end{equation}
where $\mathcal{L}_{\text{sim}}$, $\mathcal{L}_p$, and $\mathcal{L}_o$ denote the similarity matrix, positional head, and offset head losses, respectively. And $\mathcal{L}_I$ corresponds to the IHERM loss. Gradients from $\mathcal{L}_I$ are detached from the SDBM to avoid convergence issues arising from abnormal gradients caused by highly distorted inputs for IHERM in the early training stages. 

\textbf{Similarity matrix loss $\mathcal{L}_{sim}$} is computed by focal loss (FL)~\cite{lin2017focal} between predicted cross-scale similarity matrix  $\hat{S}$ and its ground truth $S_{gt}$:
\begin{equation}
\mathcal{L}_{sim} = \sum_{i,j} FL(\hat{S}(i,j), S_{gt}(i,j)),
\label{eq:sim_loss}
\end{equation}
$S_{gt}(i,j)=1$ if the point represented by $i$ is matched with the point represented by $j$. The focal loss FL is formulated as:
\begin{equation}
FL(p,y) = \begin{cases} -\alpha_t(1-p)^\gamma \log(p_t), & \text{if }  y=1,\\
-\alpha_{t}\cdot {p}^\gamma \cdot \log(p_t), & \text{if } y=0,
\end{cases}
\end{equation}
where $\alpha_{t}$ and $\gamma$ are set to $0.95$ and $2$ by default.

\textbf{Position prediction head loss $\mathcal{L}_{p}$} is computed using the predicted position map and its ground truth:
\begin{equation}
\mathcal{L}_p = \frac{1}{KN}\sum_{i=1}^{K} \sum_{x,y} FL(\mathbf{HM}(i,x,y), \mathbf{HM}_{gt}(i,x,y)),
\end{equation}
where $N=H_s \times W_s$, $K$ is the number of grid-sampled keypoints, and $\mathbf{HM}_{gt}$ is formalized as follows:
\begin{equation}
\mathbf{HM}_{gt}(i, x, y) = 
\begin{cases} 
1, & \text{if } |x - \frac{X_{i}^{gt}}{2^k}| \leq r_p \text{,} |y - \frac{Y_{i}^{gt}}{2^k}| \leq r_p \\
0, & \text{otherwise}
\end{cases},
\end{equation}
where $r_p$ (1 as default) is the positive range, $2^k$ is the downsampling factor. $(X_{i}^{gt},Y_{i}^{gt})$ is the ground truth for the $i$-th predicted keypoint.

\textbf{Offset prediction head loss $\mathcal{L}_{o}$} is calculated using the Generalized Intersection over Union (GIoU)~\cite{rezatofighi2019generalized} with box size $h_s\times w_s$:
\begin{equation}
\mathcal{L}_o =  \sum_{i=1}^{K} \frac{1}{K|\mathcal{P}_i|}\sum_{(x_i^m, y_i^m) \in \mathcal{P}_i} 1-\text{GIoU}(F_n(x_i^m,y_i^m), (X_{i}^{gt},Y_{i}^{gt})),
\end{equation}
where $\mathcal{P}_i = \{(x_i^m, y_i^m) | \mathbf{HM}_{gt}(i, x_i^m, y_i^m) = 1\}$ is the set of positive points in $\mathbf{HM}_{gt}$, and $F_n$ is defined in Equation~(\ref{eq:F_n}).

\textbf{IHERM loss} $\mathcal{L}_{I}$ is formulated as:
\begin{equation}
\mathcal{L}_{I}=\sum_{q=1}^{3Q}\left(\|\mathbf{T}_q-\mathbf{T}_{\mathrm{gt}}\|_1+\mathcal{L}_{\mathrm{FGO}}(\|\mathbf{T}_q-\mathbf{T}_{\mathrm{gt}}\|_1)\right),
\end{equation}
where $Q$ denotes the number of iterations on each scale, $T_q$ the accumulated offset at iteration $q$, and $T_{gt}$ the ground truth offset. The Fine-Grained Optimization loss $\mathcal{L}_{\mathrm{FGO}}$~\cite{zhu2024mcnet} is defined as:
\begin{equation}
\mathcal{L}_{\mathrm{FGO}}(t) = 
\left\{
\begin{array}{ll}
0 & \quad t \geq \beta, \\
-\dfrac{1}{t + \epsilon} & \quad t < \beta,
\end{array}
\right.
\end{equation}
where $\beta$ and $\epsilon$ are set to 0.85 and 0.1 by default.

\begin{figure}[t]
    \centering
    \includegraphics[width=1\columnwidth]
    {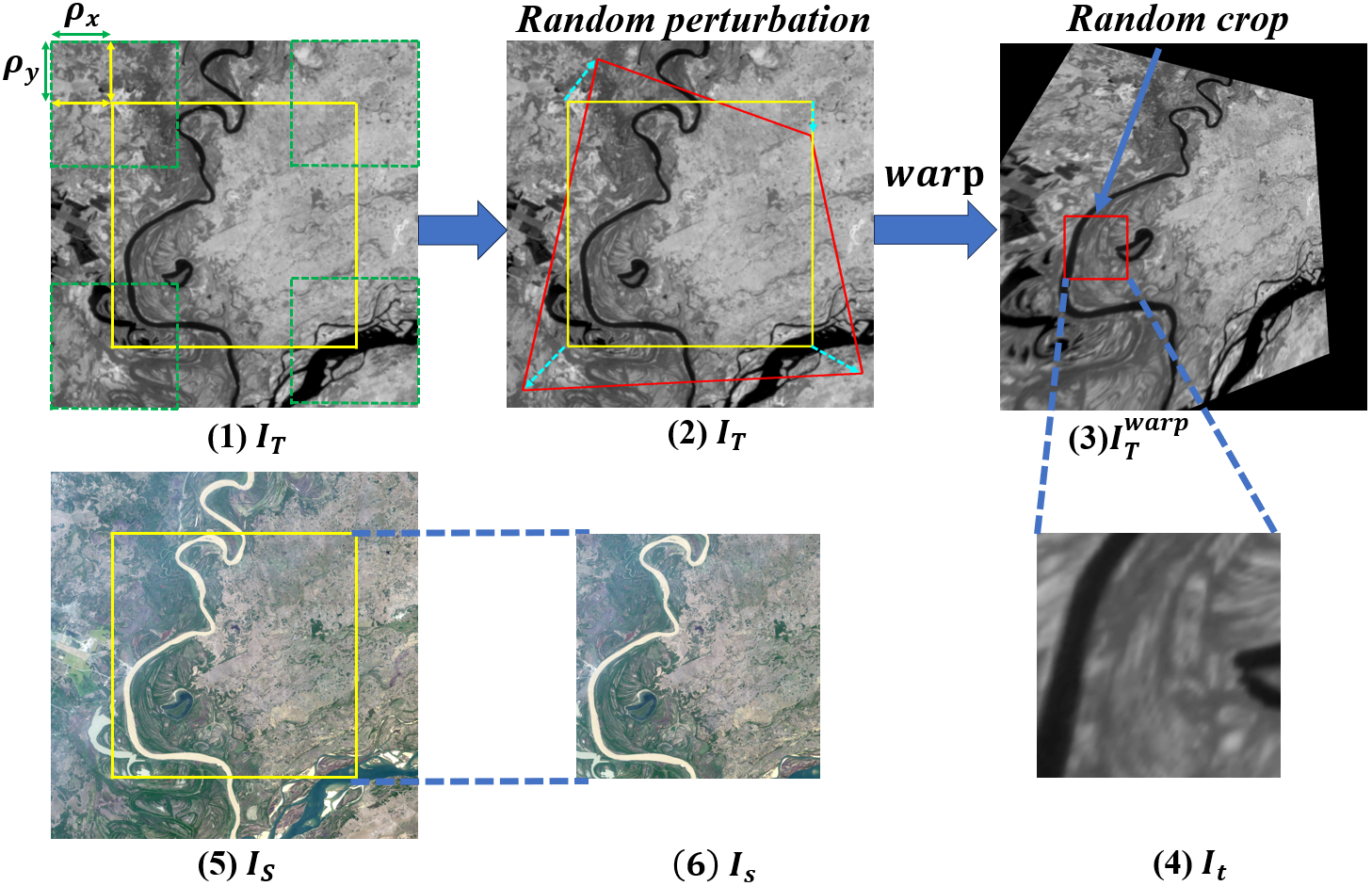}
    \caption{Overview of the data generation pipeline. }
    \label{fig:data_gen}
\end{figure}

\begin{figure*}[t]
    \centering
    \includegraphics[width=1\textwidth]{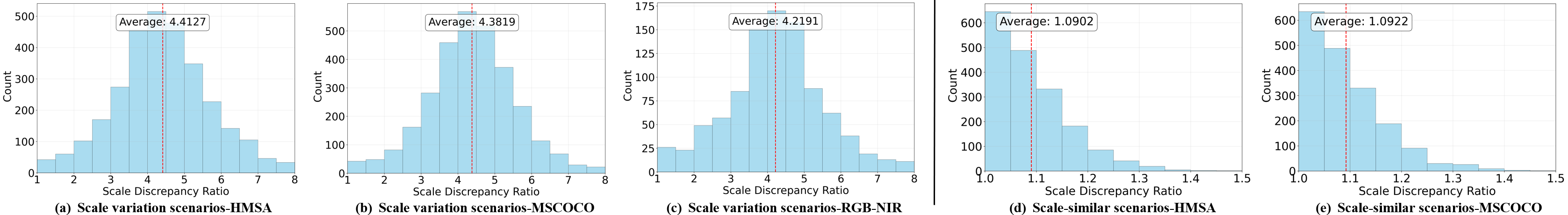}
    \caption{Scale Discrepancy Ratio (SDR) distributions of the validation datasets. (a)–(c) illustrate the SDR distributions of the HMSA, MSCOCO~\cite{lin2014microsoft}, and RGB-NIR~\cite{brown2011multi} validation sets under scale variation scenarios, respectively. (d)–(e) present the SDR distributions of the HMSA and MSCOCO~\cite{lin2014microsoft} validation sets under conventional scale-similar scenarios.}
    \label{fig:validation datasets scale discrepancy distribution}
\end{figure*}

\subsection{Data Generation}
\label{sec:data_gen}
Our data generation pipeline is illustrated in Fig.~\ref{fig:data_gen}. Given aligned image pairs $I_S$ and $I_T$, we denote $p_{org}$ as the four corners of the patch with size $H_t\times W_t$ in $I_S$. First, we perturb $p_{org}$ by adding x-coordinate offsets randomly sampled from $(-\rho_x, \rho_x)$ and y-coordinate offsets randomly sampled from $(-\rho_y, \rho_y)$ to generate $p_{dst}$. Next, we compute homography $H_g$ mapping $p_{org}$ to $p_{dst}$, and apply it to warp $I_T$, yielding $I_T^{warp}$. Subsequently, we randomly crop a patch $I_T^{p}$ of size $(H_t / z, W_t / z)$ from $I_T^{warp}$, where $z$ is uniformly sampled from the interval $[1, Z]$. We then resize the patch $I_T^{p}$ to dimensions $(H_t, W_t)$ to form $I_t$. Finally, we crop the patch of size $H_s\times W_s$ from $I_S$ at the location corresponding to $p_{org}$. Thus, our training data pair $(I_t, I_s)$ is constructed. The scale discrepancy ratio is calculated as follows:

\begin{equation}
SDR = \max \left( \sqrt{\frac{\text{area}(I_t^{back})}{\text{area}(I_s)}}, \sqrt{\frac{\text{area}(I_s)}{\text{area}(I_t^{back})}} \right).
\label{eq:sdr_final}
\end{equation}
where $I_t^{back}$ is the corresponding region of $I_t$ in $I_T$. The SDR reflects the approximate discrepancy ratio of the spatial distance represented by a single pixel between $I_s$ and $I_t$. When the overlap area between $I_t$ and $I_s$ is less than half of the area of $I_t^{back}$, the crop process for generating $I_t$ is restarted.

\section{Experiments}
\label{experiments}
\subsection{Datasets}
We evaluate our method on both uni-modal and cross-modal datasets under two scenarios: the challenging scale variation scenario and the traditional scale-similar scenario. In the scale variation scenario, we validate our method on the proposed HMSA dataset, MSCOCO~\cite{lin2014microsoft}, and RGB-NIR scene~\cite{brown2011multi}. In the traditional scale-similar scenario, we utilize MSCOCO~\cite{lin2014microsoft}, DroneVehicle~\cite{sun2022drone}, HMSA, GoogleMap~\cite{zhao2021deep}, and GoogleEarth~\cite{zhao2021deep}. The details of the datasets are as follows:

\indent 1) \textbf{HMSA dataset}: We collected and constructed a dataset named HMSA (High-resolution, Multi-modal, Satellite, Aligned dataset) from publicly available Landsat8 satellite imagery. HMSA comprises 11,837 registered visible-infrared satellite image pairs with $1152\times1152$ resolution, comprising 10,478 pairs for training and 1,359 for testing. It represents a cross-modal aerial type dataset where homography estimation under scale variations is frequently required. 

\indent 2) \textbf{MSCOCO dataset}: MSCOCO~\cite{lin2014microsoft} serves as a standard uni-modal benchmark and is widely adopted for validating homography estimation methods in previous works~\cite{detone2016deep,shao2021localtrans,cao2022iterative,cao2023recurrent,zhu2024mcnet,zhang2025adapting}. The dataset comprises 118,287 RGB images for training and 40,670 for testing.

\indent 3) \textbf{RGB-NIR dataset}: Commonly adopted as benchmark in recent works~\cite{deng2024crosshomo,zhang2025adapting}, this dataset features paired RGB and Near-Infrared (NIR) images, with 400 registered image pairs for training and 77 registered image pairs for testing.

\indent 4) \textbf{DroneVehicle dataset}: This is a challenging cross-modal dataset containing visible-infrared (VIS-IR) image pairs of traffic scenes captured at night, exhibiting significant modal discrepancies. In particular, consistent with the generation algorithm and the data split configuration in~\cite{zhang2025adapting}, we utilize 10,000 registered image pairs for training and 1,000 pairs images generated by~\cite{zhang2025adapting} for testing.

\indent 5) \textbf{GoogleMap and GoogleEarth datasets}: Following previous works~\cite{shao2021localtrans,cao2022iterative,cao2023recurrent,zhu2024mcnet}, we conduct experiments on these two widely-used datasets. The GoogleMap~\cite{zhao2021deep} dataset consists of aligned image pairs of aerial photographs and corresponding maps, comprising 8,822 training pairs and 888 test pairs. The GoogleEarth~\cite{zhao2021deep} dataset comprises aligned image pairs captured at different times from the same locations, containing 8,750 training pairs and 850 validation pairs. For GoogleEarth~\cite{zhao2021deep} and GoogleMap~\cite{zhao2021deep} datasets, we adopt the data generation algorithm and the validation sets as established in prior works~\cite{cao2022iterative,cao2023recurrent,zhu2024mcnet}.

Evaluation benchmarks were constructed for HMSA, MSCOCO~\cite{lin2014microsoft}, and RGB-NIR~\cite{brown2011multi} by sampling from their respective validation splits and applying the data generation pipeline detailed in Sec.~\ref{sec:data_gen}. Since randomly cropped regions may be textureless and uninformative for estimation, we filter out such instances, as well as those where the Scale Discrepancy Ratio (SDR) exceeds 8. The parameters for generation are detailed in Table~\ref{tab:benchmark_config}, and the statistical distribution of the SDR across the constructed benchmarks is presented in Fig.~\ref{fig:validation datasets scale discrepancy distribution}. 

\begin{table}[h]
    \centering
    \caption{Configuration for Evaluation Benchmark Generation.}
    \label{tab:benchmark_config}
    \begin{tabular}{llll}
        \toprule
        \textbf{Scenario} & \textbf{Dataset} & \textbf{Size} & \textbf{Params ($\rho, Z$)} \\
        \midrule
        \multirow{3}{*}{\textbf{Scale Variation}} 
        & HMSA & 3,000 & $\rho_x=\rho_y=192, Z=5$ \\
        & MSCOCO & 3,000 & $\rho_x=\rho_y=192, Z=5$ \\
        & RGB-NIR & 950 & $\rho_x=\rho_y=192, Z=5$ \\
        \midrule
        \multirow{3}{*}{\textbf{Scale-Similar}} 
        & HMSA & 1,800 & $\rho_x=\rho_y=192, Z=1$ \\
        & MSCOCO & 1,800 & $\rho_x=\rho_y=192, Z=1$ \\
        \bottomrule
    \end{tabular}
\end{table}

\begin{table*}[h] 
\centering
\caption{Performance comparison on scale variation scenarios using MACE. We report model performance across three scale discrepancy ratio ranges: Small Group (1.0-1.5), Medium Group (1.5-4.0), and Large Group (4.0-8.0).}
\label{tab:performance_scale} 
\resizebox{\textwidth}{!}{ 
    \begin{tabular}{l|cccc|cccc|cccc}
    \toprule
    \multirow{2}{*}{Method}  & \multicolumn{4}{c|}{MSCOCO ($768\times 768$)} & \multicolumn{4}{c|}{HMSA ($768\times 768$)} & \multicolumn{4}{c}{RGB-NIR ($768\times 768$)} \\
    \cmidrule(lr){2-5} \cmidrule(lr){6-9} \cmidrule(l){10-13}
    &  Small$ \downarrow$ & Medium$ \downarrow$ & Large$ \downarrow$ & Avg $\downarrow$ & Small $\downarrow$& Medium $\downarrow$& Large $\downarrow$ & Avg $\downarrow$ & Small $\downarrow$& Medium $\downarrow$& Large $\downarrow$ & Avg $\downarrow$ \\
    \midrule
    SIFT+RANSAC~\cite{lowe2004distinctive}  & 0.38 & 70.77 & 173.43 & 135.66 & 439.57 & 523.73 & 540.72 & 533.28  & 23.66 &252.96& 327.08 & 290.30 \\
    DHN~\cite{detone2016deep}  & 32.51 & 58.45 & 81.26 & 72.72 & 128.75 & 235.94 & 236.93 & 235.06 & 47.98 & 107.66 & 132.49 & 120.64 \\
    Localtrans~\cite{zhao2021image} & 9.85 & 47.08 & 90.55 & 74.46 & 131.77 & 254.16 & 245.82 & 247.18  & 38.82 & 77.81 & 122.97 & 103.31 \\
    IHN~\cite{cao2022iterative}  & 1.85 & 27.16 & 79.25 &  60.23 & 154.57 & 289.05 & 264.87 & 271.90 & 5.07 & 55.24 & 104.17 & 82.66 \\
    RHWF~\cite{cao2023recurrent}  & 0.30  & 26.08 & 72.07 & 55.23 & 152.64  & 292.02 & 264.63 & 272.77  & 4.48 & 54.56 & 97.12 & 78.23 \\
    McNet~\cite{zhu2024mcnet} & 3.44  & 44.45 & 100.16 & 79.62 & 124.64  & 274.83 & 263.10 & 265.32  & 7.70 & 72.89 & 111.90 & 94.06 \\
    P2WNet~\cite{xie2025p2wnet} & 157.75  & 24.73 & \underline{15.38} & 20.59 & 240.98  & 100.99 & \underline{60.51} & \underline{77.38}  & 194.99 & 35.98 & \underline{29.37} & 36.45 \\
    GFNet~\cite{zhang2025adapting}  & \underline{0.28}  & \underline{2.57} & 18.16 & \underline{12.54} & \underline{0.80}  & \underline{82.13} & 129.11 & 110.67  & 
    \underline{2.15} & \underline{15.82} & 34.56 & \underline{26.47} \\
    \midrule
    SA-Homo (Ours)  & \textbf{0.028}  & \textbf{0.032} & \textbf{0.046} & \textbf{0.041} & \textbf{0.414} & \textbf{0.298} & \textbf{0.303} & \textbf{0.304} & \textbf{2.384}& \textbf{1.809}& \textbf{2.167}& \textbf{2.036} \\
    \bottomrule
    \end{tabular}
}
\end{table*}

\begin{figure*}[h]
    \centering
    \includegraphics[width=1\textwidth]{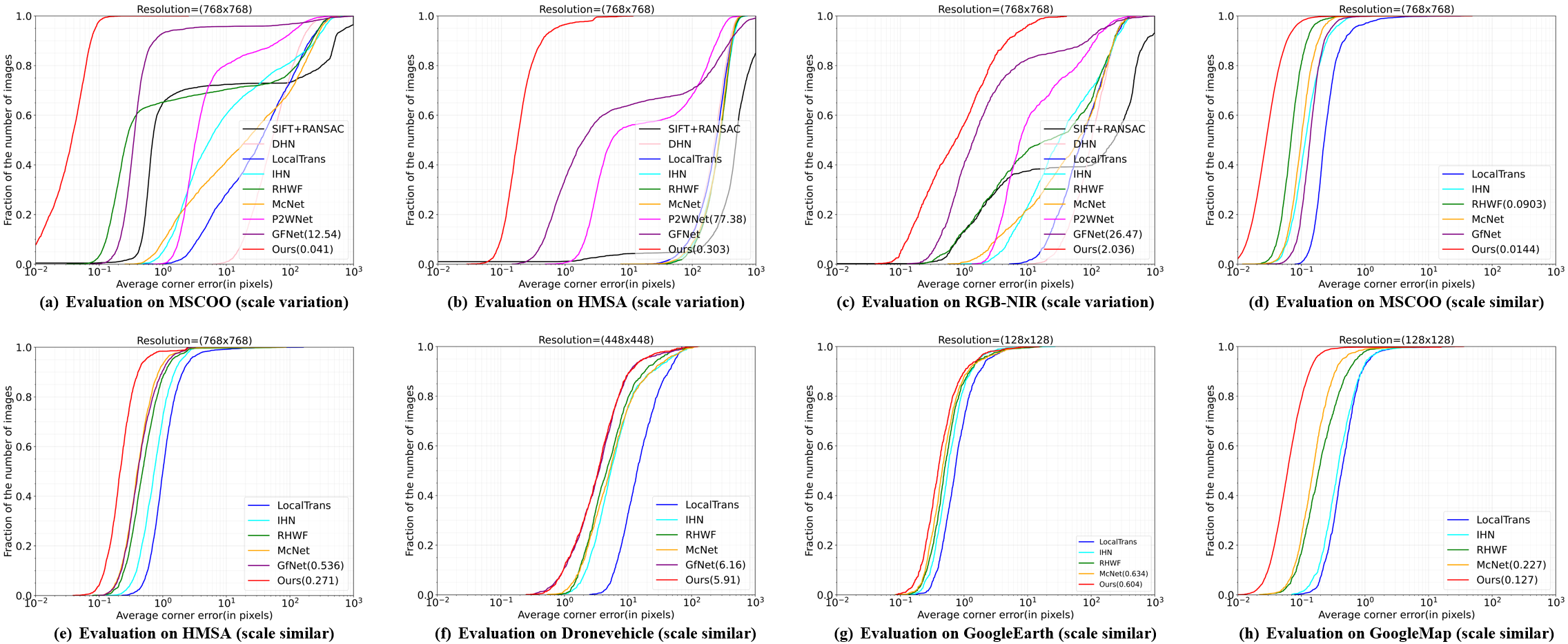}
    \caption{Performance comparison under two settings: the scale variation scenario on (a) MSCOCO~\cite{lin2014microsoft}, (b) HMSA, and (c) RGB-NIR~\cite{brown2011multi}; and the conventional scale-similar scenario on (d) MSCOCO~\cite{lin2014microsoft}, (e) HMSA, (f) DroneVehicle~\cite{sun2022drone}, (g) GoogleEarth~\cite{zhao2021deep} and (h) GoogleMap~\cite{zhao2021deep}. The x-axis represents the estimated average corner error (ACE), and the y-axis represents the fraction of data below the corresponding ACE. We present the MACE of the best-performing and second-best-performing methods in the figure.}
    \label{fig:CDF}
\end{figure*}

\begin{figure*}[t]
    \centering
    \includegraphics[width=1\textwidth]{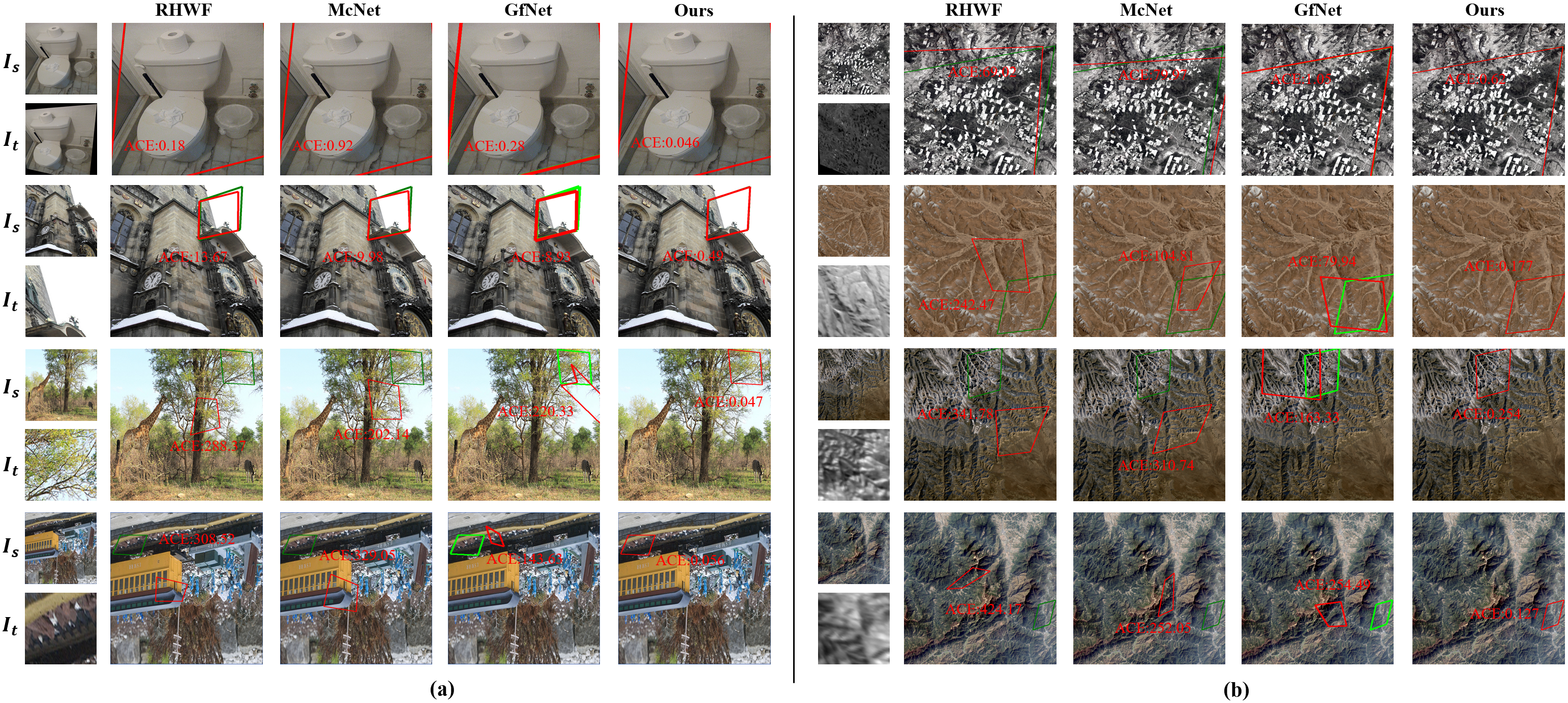}
    \caption{The Visualization results on (a) MSCOCO dataset~\cite{lin2014microsoft}, (b) HMSA dataset. Red and green bounding boxes indicate the prediction and ground truth respectively. In (a), we observe that as the scale discrepancy between image pairs increases progressively (from top to bottom rows), the estimation accuracy of previous method deteriorates rapidly, while our method maintains stable estimation performance across varying scale differences. In (b), we demonstrate that our model maintains robust estimation performance under scale variations and cross-modality scenarios. }
    \label{fig:visualization on scale variation scenarios}
\end{figure*}

\begin{figure*}[h]
    \centering
    \includegraphics[width=1\textwidth]{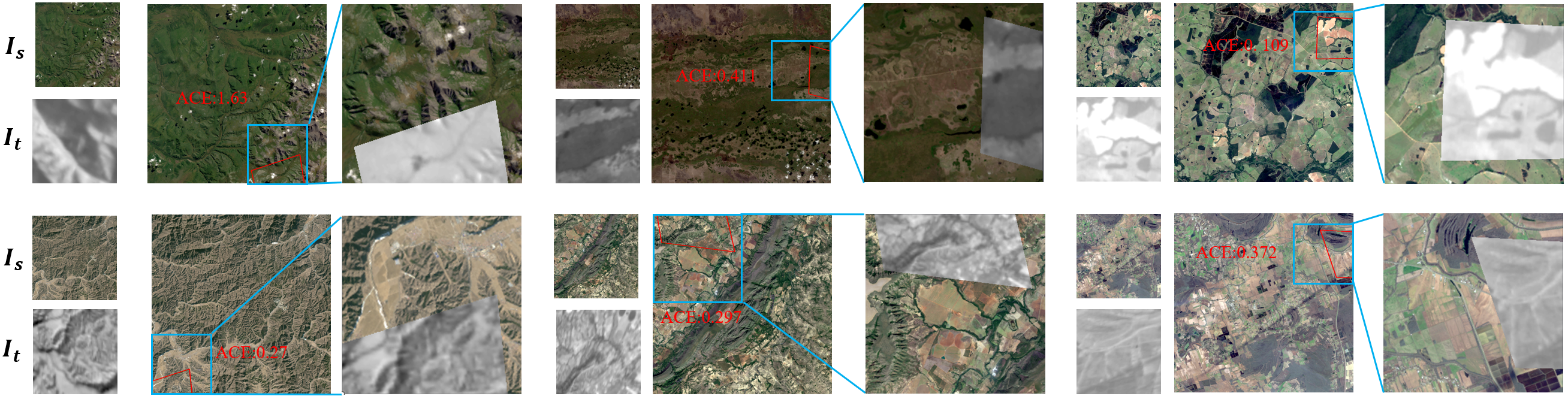}
    \caption{Visualization of boundary cases, where only part of the $I_t$ lies within the $I_s$. The red and green bounding boxes denote the prediction and ground truth, respectively, which almost overlap. The blue box indicates the zoom-in area. The template image is mapped back to the search image using the estimated  $\mathbf{H}$, and the mapped content replaces the original region.
    }
    \label{fig:boundary cases}
\end{figure*}

\subsection{Experimental Settings}
\textbf{Implementation Details}. Our method is implemented in PyTorch and trained using the AdamW optimizer~\cite{loshchilov2017decoupled} with a batch size of 8 and a learning rate of $2\times 10^{-4}$. The training process consists of 120,000 iterations for the MSCOCO~\cite{lin2014microsoft} dataset and 500,000 iterations for the HMSA, RGB-NIR~\cite{brown2011multi}, DroneVehicle~\cite{sun2022drone}, GoogleMap~\cite{lin2014microsoft}, GoogleEarth~\cite{lin2014microsoft}. All experiments were conducted using two NVIDIA RTX 3090 GPUs and Intel(R) Xeon(R) Gold 5320 CPUs.

The dimensions of $I_t$ and $I_s$ were set to $768 \times 768$ for MSCOCO, HMSA, and RGB-NIR~\cite{brown2011multi}, $448 \times 448$ for DroneVehicle~\cite{sun2022drone}, and $128 \times 128$ for GoogleMap and GoogleEarth~\cite{zhao2021deep}. 
For HMSA, MSCOCO, and RGB-NIR~\cite{brown2011multi} datasets, we employed the generation algorithm described in Sec.~\ref{sec:data_gen} to construct the training data, the perturbation range $(-\rho_y, \rho_y)$ was set to $(192, 192)$, the parameter $Z$ was set to 4 for scale-variant scenarios and 1 for scale-similar scenarios. Furthermore, to comprehensively evaluate the performance of our method under conventional scale-similar settings, we adopted the data generation strategy proposed in~\cite{zhang2025adapting} for the DroneVehicle~\cite{sun2022drone}, and the generation strategy used in~\cite{cao2022iterative,cao2023recurrent,zhu2024mcnet} for the GoogleMap and GoogleEarth~\cite{zhao2021deep}.

The original networks in~\cite{zhao2021image,cao2022iterative,cao2023recurrent,zhu2024mcnet} were designed for input images with a resolution of $128 \times 128$. For experiments with larger input resolution, we adapted the network configurations to support the larger input image pairs.

In our optimal setup, the model comprises $N_c = 2$ attention layers  with $D=256$ channel dimension in MLAC, $T=5,\tau=1$ when applying Sinkhorn–Knopp algorithm, $K = 144$ grid-sampled keypoints in SDBM, and $Q = 2$ iterations per scale in IHERM. Both the feature extractor and the IHERM module utilize 96 channels. Regarding the loss function, the hyperparameters are set as follows: $\alpha = 10^{-3}$, $\alpha_t = 0.95$, $\gamma = 2$, $\beta = 0.85$, and $\epsilon = 0.1$, respectively. RANSAC~\cite{fischler1981random} is applied using the default settings of the \texttt{findHomography.cv2} in OpenCV. We employ iteration on downsampling scale sequences of $\{4, 2, 1\}$ and $\{16, 8, 4\}$ for inputs resolution of $128 \times 128$ and $768 \times 768$, respectively. For the $448 \times 448$  dataset, images are resized to $768 \times 768$ for both training and inference, and the predictions are rescaled back to the original $448 \times 448$ resolution for error evaluation.

\subsection{Evaluation}
We compare our method against the latest homography estimation methods, including GFNet~\cite{zhang2025adapting}, McNet~\cite{zhu2024mcnet}, RHWF~\cite{cao2023recurrent}, IHN~\cite{cao2022iterative}, and LocalTrans~\cite{shao2021localtrans}, as well as the traditional methods DHN~\cite{detone2016deep} and SIFT+RANSAC. We adopt Mean Average Corner Error (MACE) as the evaluation metric, following previous work~\cite{zhu2024mcnet,cao2023recurrent,cao2022iterative,shao2021localtrans,zhang2025adapting}. MACE is computed as follows (truncated at 1000):
\begin{equation}
\mathrm{MACE}
= \frac{1}{N}\sum_{i=1}^{N}\frac{1}{4}\sum_{j=1}^{4}
\left\lVert \hat{\mathbf{c}}_{i,j}-\mathbf{c}_{i,j}\right\rVert_{2},
\label{eq:mace}
\end{equation}
where \(N\) is the total number of image pairs, \(\hat{\mathbf{c}}_{i,j}\) denotes the \(j\)-th predicted corner position of the \(i\)-th image pair, and \(\mathbf{c}_{i,j}\) represents the corresponding ground truth corner position.

\begin{figure*}[h]
    \centering
    \includegraphics[width=1\textwidth]{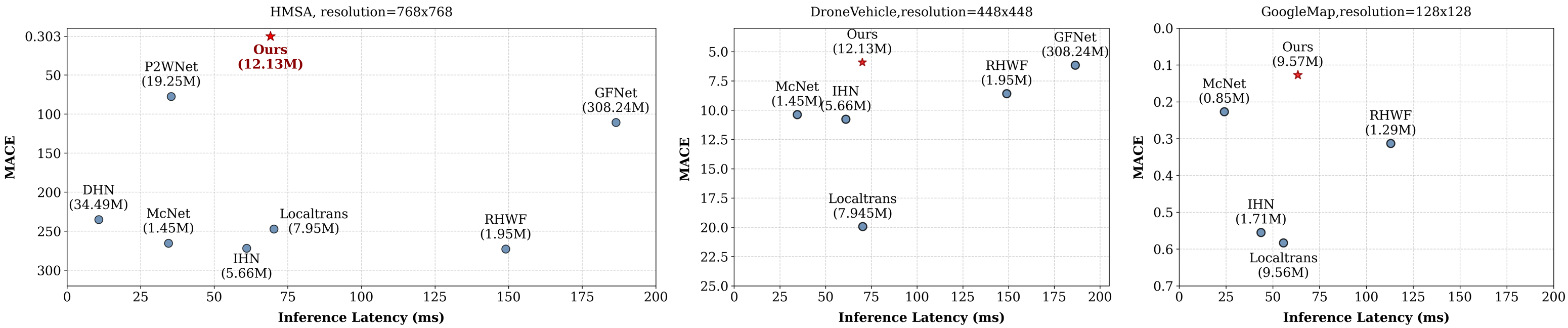}
    \caption{Runtime and efficiency analysis. We present the inference latency, total parameters, and MACE of state-of-the-art deep homography estimation methods on the HMSA (scale variations), MSCOCO~\cite{lin2014microsoft}, and GoogleMap~\cite{zhao2021deep} (scales-similar) datasets. All evaluations were performed on an NVIDIA RTX 3090 GPU with a batch size of 1.}
    \label{fig:mace_vis_time}
\end{figure*}

\textbf{Evaluation on scale variation scenarios.} As shown in Table~\ref{tab:performance_scale}, we partition the image pairs in the validation set into three groups according to the scale discrepancy ratio: small (1–1.5), medium (1.5–4), and large (4–8). We present the average MACE of the latest deep homography estimation methods for each group. It can be observed that on the uni-modal MSCOCO~\cite{lin2014microsoft} dataset, previous methods suffer from performance degradation as the scale discrepancy varies. This degradation becomes even worse when they face both cross-modal and scale variations (Fig.~\ref{fig:mace_vs_sdr} (b)-(c)). In contrast, our model shows superior estimation accuracy and scale robustness, maintaining consistent performance across different scale discrepancy ratio groups (shown in Table~\ref{tab:performance_scale}). Furthermore, our method achieves MACE of 0.304 and 1.891 on the HMSA and RGB-NIR~\cite{brown2011multi} datasets, respectively, demonstrating its effectiveness in cross-modal homography estimation under scale variation scenarios. We further evaluate performance using Cumulative Distribution Function (CDF) plots (Fig.~\ref{fig:CDF} (a)--(c)), where our method's curves are positioned closest to the top-left corner across all three datasets, demonstrating superior performance. Beyond quantitative metrics, we provide visualization results in Fig.~\ref{fig:visualization on scale variation scenarios}. In practical scenarios, the template $I_t$ is not always fully contained within the search image $I_s$. Therefore, we explicitly evaluate the robustness of our method in boundary cases, as shown in Fig.~\ref{fig:boundary cases}. The results demonstrate that our method effectively handles image pairs where $I_t$ only partially overlaps with $I_s$.

\textbf{Evaluation on conventional scale-similar scenarios.} As shown in Fig.~\ref{fig:CDF} (e)--(h), we evaluate the performance of our model under conventional scale-similar scenarios. Our method surpasses the performance of previous approaches across different datasets. For the cross-modal HMSA dataset, our method achieves a $49.44\%$ (Fig.~\ref{fig:CDF} (e)) improvement over the SOTA method GFNet~\cite{zhang2025adapting}. For the DroneVehicle dataset, which is consistent with~\cite{zhang2025adapting}, our method also demonstrates slightly better performance than GFNet. Since the models in previous works~\cite{zhao2021image,cao2022iterative,cao2023recurrent,zhu2024mcnet} were specifically designed for $128\times128$ resolution image inputs, we also conducted performance comparisons on the GoogleMap and GoogleEarth datasets utilized in~\cite{zhao2021image,cao2022iterative,cao2023recurrent,zhu2024mcnet} with $128\times128$ resolution. It can be observed that our method achieves a $44.05\%$ (Fig.~\ref{fig:CDF} (h)) performance improvement over the previous SOTA method McNet~\cite{zhu2024mcnet} on GoogleMap. These experiments indicate that the hierarchical scale alignment strategy adopted by our method also improves performance in conventional scale-similar scenarios. This can be attributed to the clear division of responsibilities: SDBM handles scale-induced discrepancies (even when they are less pronounced in scale-similar settings), while IHERM focuses more on fine-grained refinement under nearly aligned conditions.

\textbf{Computational analysis.} We compare the accuracy, inference latency including RANSAC overhead, and number of parameters of different deep homography estimation methods in Fig.~\ref{fig:mace_vis_time}. Our method achieves favorable accuracy compared with recent strong baselines such as GFNet~\cite{zhang2025adapting} and RHWF~\cite{cao2023recurrent}, while maintaining lower inference latency. These results suggest that the proposed design is effective in handling large-scale variations and provides a balanced trade-off between estimation performance and computational efficiency. This efficiency can be partly attributed to the heavy-to-lightweight module design.

\begin{figure*}[t]
    \centering
    \includegraphics[width=1\textwidth]{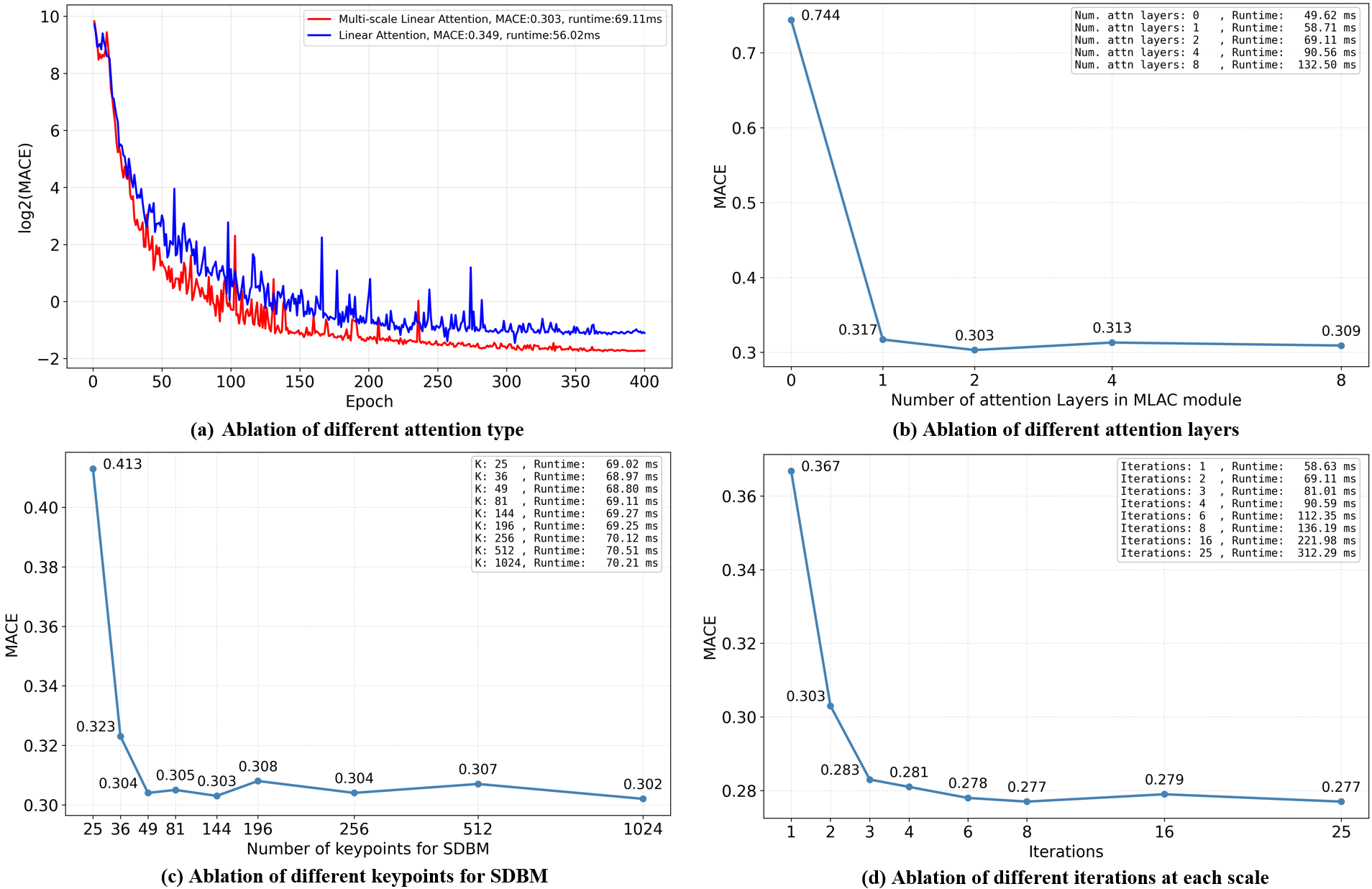}
    \caption{Ablation studies on the HMSA dataset. We investigate the impact of: (a) attention mechanisms (linear vs. multi-scale) in MLAC; (b) the number of stacked attention layers in MLAC; (c) the quantity of keypoints in SDBM; and (d) the iteration counts per scale in the IHERM refinement process.}
    \label{fig:Ablation_study_vis}
\end{figure*}

\begin{table}[H]
\caption{Ablation study of our method on HMSA dataset.}
\centering
\begin{tabular}{l|c|c|c}
\toprule
 & Setting  & MACE$\downarrow$ &  Time(ms)$\downarrow$\\
\midrule
\multirow{3}{*}{(a) Main Module} & only SDBM  & 1.945 & 36.62 \\
& only IHERM & 255.31 & 31.49 \\
& \textbf{both} & 0.303 & 69.11 \\
\midrule
\midrule
\multirow{2}{*}{(b) Robust Estimation} & w/o RANSAC  & 56.13 & 65.22 \\
& \textbf{w/ ~RANSAC} & 0.303 & 69.11 \\
\midrule
\multirow{4}{*}{(b) Iteration scale} & [16-2] & 0.776 & 47.99 \\
& [16,8-2,2] & 0.526 & 56.78 \\
& \textbf{[16,8,4-2,2,2]} & 0.303 & 69.11 \\
& [16,8,4,2-2,2,2,2] & 0.209 & 120.27 \\
\bottomrule
\end{tabular}
\label{tab:Ablation_study}
\end{table}

\subsection{Ablation Study}
We conduct ablation studies on the HMSA dataset under scale variation scenarios to evaluate the effectiveness of each main component and the impact of hyperparameter selections. We use the best practice configuration as the baseline.

\textbf{Main Module.} As shown in Table~\ref{tab:Ablation_study} (a), we validate the individual contributions of the SDBM and IHERM modules to demonstrate the effectiveness of our hierarchical scale alignment strategy. The results show that the SDBM module exhibits the capability to perform homography estimation under scale variation scenarios. In contrast, the IHERM module alone fails to produce reliable homography estimations when confronted with scale variations, highlighting the critical role of both SDBM and IHERM. It can be observed that in an iterative module, significant estimation errors in the initial iteration will distort the image and substantially impair the predictions of subsequent iterations. Consequently, a capable SDBM module with a design for handling large scale discrepancy image pairs is necessary for initial alignment under scale variation scenario.

\textbf{Robust Estimation.} Table~\ref{tab:Ablation_study} (b) demonstrates the necessity of using RANSAC~\cite{fischler1981random} to remove unreliable correspondences. Since our loss function does not impose supervision on template keypoints whose corresponding locations fall outside the search image, such cases may introduce noisy matches during inference. In addition, correspondence establishment is more challenging and further increase the likelihood of outliers under scale variation scenarios. Consequently, RANSAC is indispensable for outlier rejection and contributes to more robust initial alignment.

\textbf{Multi-scale Linear Attention.} As illustrated in Fig.~\ref{fig:Ablation_study_vis} (a), by incorporating Multi-scale Linear Attention~\cite{cai2022efficientvit}, we achieve a performance improvement of 15.18\% over standard Linear Attention~\cite{katharopoulos2020transformers}. This gain is attributed to the effective fusion of multi-scale information, which is important for achieving robust initial alignment in scale variation scenarios.

\begin{table}[t]
\caption{Ablation study of CSMB module on HMSA dataset. }
\centering
\begin{tabular}{c|c|c|c}
\toprule
Setting &SDBM MACE$\downarrow$ &MACE$\downarrow$& Time(ms)$\downarrow$\\
\midrule
\textbf{ w/ both } & 1.95& 0.303 & 69.11 \\
w/o multi-scale &2.31& 0.313 & 68.62 \\
w/o sinkhorn-Knopp & 2.56 & 0.321 & 67.79 \\
w/o both & 2.73& 0.331 & 65.82 \\
\bottomrule
\end{tabular}
\label{tab:Ablation_study_scem}
\end{table}

\textbf{Components in CSMB.} Table~\ref{tab:Ablation_study_scem} (b) demonstrates that both the multi-scale design and Sinkhorn-Knopp algorithm\cite{cuturi2013sinkhorn} in proposed CSMB provide modest gains in final MACE at the expense of a minor computational overhead. Their primary value lies in facilitating robust initial alignment. The employment of both the Sinkhorn-Knopp algorithm and the multi-scale design reduces the MACE of SDBM from 2.73 to 1.95. We attribute the performance gains of the multi-scale design in CSMB to its effective aggregation of correlation representations across multi-granularity features, which alleviates local feature inconsistencies caused by scale variations. The application of the Sinkhorn-Knopp algorithm to the similarity matrix mitigates many-to-one matching ambiguities, thereby boosting the performance of our initial alignment.

\textbf{Attention layers and number of keypoints.} As illustrated in Fig.~\ref{fig:Ablation_study_vis} (b), for the MLAC module, stacking more than 2 attention layers does not result in continuous performance improvement but introduces significant computational overhead. Consequently, we adopt $N_c=2$ attention layers. Similarly, as shown in  Fig.~\ref{fig:Ablation_study_vis} (c), we observe that in the SDBM module, increasing the number of grid-sampled keypoints beyond 49 yields no further significant performance gains. We set $K=144$ grid-sampled keypoints for SDBM prediction. An excessive number of keypoints would require more time for RANSAC~\cite{fischler1981random} during the early training stage, when correspondences are ambiguous.
 
\textbf{Iteration Scales and Counts.} 
As reported in Table~\ref{tab:Ablation_study} (c), we observe that incorporating iterative refinement at finer-scale feature maps can enhance the estimation accuracy. Specifically, extending the iterative process to encompass the $\{16, 8, 4, 2\}$ downsampling scales yields a substantial performance gain of $31.0\%$ over the $\{16, 8, 4\}$ configuration. However, this improvement comes at the expense of a substantial increase in inference latency, which rises from 69.11 ms to 120.27 ms. Furthermore, as illustrated in Fig.~\ref{fig:Ablation_study_vis} (d), we validate that increasing the number of iterations per scale beyond 2 results in diminishing returns, yielding only marginal performance gains while incurring considerable computational overhead. Consequently, to achieve an optimal balance between performance and efficiency, we adopt the $\{16,8,4\}$ downsampling scales for iterative refinement, with 2 iterations per scale.
\section{Conclusion}
\label{conclusion}

In this work, we present SA-Homo, a scale-adaptive framework designed to tackle homography estimation in challenging scale variation scenarios where existing homography estimation methods show limitation. By leveraging a hierarchical scale alignment strategy, alongside a tailored SDBM module for robust initial alignment and an IHERM module for iterative refinement, our method achieves robustness against scale variation scenarios and computational efficiency. Extensive experiments demonstrate that our method yields superior estimation accuracy in both conventional scale-similar scenarios and challenging scale-variation scenarios, remaining robust even with scale discrepancies reaching $8\times$. Finally, we release the HMSA dataset to facilitate research on homography estimation in scale variation scenarios.

\bibliographystyle{IEEEtran}
\bibliography{refs}

@inproceedings{zhan2022homography,
  title={Homography decomposition networks for planar object tracking},
  author={Zhan, Xinrui and Liu, Yueran and Zhu, Jianke and Li, Yang},
  booktitle={Proceedings of the AAAI Conference on Artificial Intelligence},
  volume={36},
  number={3},
  pages={3234--3242},
  year={2022}
}

@inproceedings{bay2006surf,
  title={Surf: Speeded up robust features},
  author={Bay, Herbert and Tuytelaars, Tinne and Van Gool, Luc},
  booktitle={Computer Vision--ECCV 2006: 9th European Conference on Computer Vision, Graz, Austria, May 7-13, 2006. Proceedings, Part I 9},
  pages={404--417},
  year={2006},
  organization={Springer}
}

@article{detone2016deep,
  title={Deep image homography estimation},
  author={DeTone, Daniel and Malisiewicz, Tomasz and Rabinovich, Andrew},
  journal={arXiv preprint arXiv:1606.03798},
  year={2016}
}

@inproceedings{erlik2017homography,
  title={Homography estimation from image pairs with hierarchical convolutional networks},
  author={Erlik Nowruzi, Farzan and Laganiere, Robert and Japkowicz, Nathalie},
  booktitle={Proceedings of the IEEE international conference on computer vision workshops},
  pages={913--920},
  year={2017}
}

@article{zhou2019stn,
  title={STN-homography: Direct estimation of homography parameters for image pairs},
  author={Zhou, Qiang and Li, Xin},
  journal={Applied Sciences},
  volume={9},
  number={23},
  pages={5187},
  year={2019},
  publisher={MDPI}
}

@inproceedings{le2020deep,
  title={Deep homography estimation for dynamic scenes},
  author={Le, Hoang and Liu, Feng and Zhang, Shu and Agarwala, Aseem},
  booktitle={Proceedings of the IEEE/CVF conference on computer vision and pattern recognition},
  pages={7652--7661},
  year={2020}
}

@inproceedings{chang2017clkn,
  title={Clkn: Cascaded lucas-kanade networks for image alignment},
  author={Chang, Che-Han and Chou, Chun-Nan and Chang, Edward Y},
  booktitle={Proceedings of the IEEE conference on computer vision and pattern recognition},
  pages={2213--2221},
  year={2017}
}

@inproceedings{zhao2021deep,
  title={Deep lucas-kanade homography for multimodal image alignment},
  author={Zhao, Yiming and Huang, Xinming and Zhang, Ziming},
  booktitle={Proceedings of the IEEE/CVF conference on computer vision and pattern recognition},
  pages={15950--15959},
  year={2021}
}

@inproceedings{cao2022iterative,
  title={Iterative deep homography estimation},
  author={Cao, Si-Yuan and Hu, Jianxin and Sheng, Zehua and Shen, Hui-Liang},
  booktitle={Proceedings of the IEEE/CVF conference on computer vision and pattern recognition},
  pages={1879--1888},
  year={2022}
}

@inproceedings{cao2023recurrent,
  title={Recurrent homography estimation using homography-guided image warping and focus transformer},
  author={Cao, Si-Yuan and Zhang, Runmin and Luo, Lun and Yu, Beinan and Sheng, Zehua and Li, Junwei and Shen, Hui-Liang},
  booktitle={Proceedings of the IEEE/CVF Conference on Computer Vision and Pattern Recognition},
  pages={9833--9842},
  year={2023}
}

@inproceedings{zhu2024mcnet,
  title={Mcnet: Rethinking the core ingredients for accurate and efficient homography estimation},
  author={Zhu, Haokai and Cao, Si-Yuan and Hu, Jianxin and Zuo, Sitong and Yu, Beinan and Ying, Jiacheng and Li, Junwei and Shen, Hui-Liang},
  booktitle={Proceedings of the IEEE/CVF Conference on Computer Vision and Pattern Recognition},
  pages={25932--25941},
  year={2024}
}

@inproceedings{shao2021localtrans,
  title={Localtrans: A multiscale local transformer network for cross-resolution homography estimation},
  author={Shao, Ruizhi and Wu, Gaochang and Zhou, Yuemei and Fu, Ying and Fang, Lu and Liu, Yebin},
  booktitle={Proceedings of the IEEE/CVF international conference on computer vision},
  pages={14890--14899},
  year={2021}
}

@article{deng2024crosshomo,
  title={CrossHomo: Cross-modality and cross-resolution homography estimation},
  author={Deng, Xin and Liu, Enpeng and Gao, Chao and Li, Shengxi and Gu, Shuhang and Xu, Mai},
  journal={IEEE Transactions on Pattern Analysis and Machine Intelligence},
  year={2024},
  publisher={IEEE}
}

@article{fischler1981random,
  title={Random sample consensus: a paradigm for model fitting with applications to image analysis and automated cartography},
  author={Fischler, Martin A and Bolles, Robert C},
  journal={Communications of the ACM},
  volume={24},
  number={6},
  pages={381--395},
  year={1981},
  publisher={ACM New York, NY, USA}
}

@inproceedings{sun2021loftr,
  title={LoFTR: Detector-free local feature matching with transformers},
  author={Sun, Jiaming and Shen, Zehong and Wang, Yuang and Bao, Hujun and Zhou, Xiaowei},
  booktitle={Proceedings of the IEEE/CVF conference on computer vision and pattern recognition},
  pages={8922--8931},
  year={2021}
}

@article{cai2022efficientvit,
  title={Efficientvit: Multi-scale linear attention for high-resolution dense prediction},
  author={Cai, Han and Li, Junyan and Hu, Muyan and Gan, Chuang and Han, Song},
  journal={arXiv preprint arXiv:2205.14756},
  year={2022}
}

@inproceedings{katharopoulos2020transformers,
  title={Transformers are rnns: Fast autoregressive transformers with linear attention},
  author={Katharopoulos, Angelos and Vyas, Apoorv and Pappas, Nikolaos and Fleuret, Fran{\c{c}}ois},
  booktitle={International conference on machine learning},
  pages={5156--5165},
  year={2020},
  organization={PMLR}
}

@article{howard2017mobilenets,
  title={Mobilenets: Efficient convolutional neural networks for mobile vision applications},
  author={Howard, Andrew G and Zhu, Menglong and Chen, Bo and Kalenichenko, Dmitry and Wang, Weijun and Weyand, Tobias and Andreetto, Marco and Adam, Hartwig},
  journal={arXiv preprint arXiv:1704.04861},
  year={2017}
}

@misc{hartley2003multiple,
  title={Multiple view geometry in computer vision},
  author={Hartley, Richard},
  year={2003},
  publisher={Cambridge university press}
}

@inproceedings{lin2017focal,
  title={Focal loss for dense object detection},
  author={Lin, Tsung-Yi and Goyal, Priya and Girshick, Ross and He, Kaiming and Doll{\'a}r, Piotr},
  booktitle={Proceedings of the IEEE international conference on computer vision},
  pages={2980--2988},
  year={2017}
}

@inproceedings{rezatofighi2019generalized,
  title={Generalized intersection over union: A metric and a loss for bounding box regression},
  author={Rezatofighi, Hamid and Tsoi, Nathan and Gwak, JunYoung and Sadeghian, Amir and Reid, Ian and Savarese, Silvio},
  booktitle={Proceedings of the IEEE/CVF conference on computer vision and pattern recognition},
  pages={658--666},
  year={2019}
}

@article{loshchilov2017decoupled,
  title={Decoupled weight decay regularization},
  author={Loshchilov, Ilya and Hutter, Frank},
  journal={arXiv preprint arXiv:1711.05101},
  year={2017}
}

@inproceedings{lin2014microsoft,
  title={Microsoft coco: Common objects in context},
  author={Lin, Tsung-Yi and Maire, Michael and Belongie, Serge and Hays, James and Perona, Pietro and Ramanan, Deva and Doll{\'a}r, Piotr and Zitnick, C Lawrence},
  booktitle={European conference on computer vision},
  pages={740--755},
  year={2014},
  organization={Springer}
}

@article{lowe2004distinctive,
  title={Distinctive image features from scale-invariant keypoints},
  author={Lowe, David G},
  journal={International journal of computer vision},
  volume={60},
  pages={91--110},
  year={2004},
  publisher={Springer}
}

@inproceedings{rublee2011orb,
  title={ORB: An efficient alternative to SIFT or SURF},
  author={Rublee, Ethan and Rabaud, Vincent and Konolige, Kurt and Bradski, Gary},
  booktitle={2011 International conference on computer vision},
  pages={2564--2571},
  year={2011},
  organization={Ieee}
}

@article{cuturi2013sinkhorn,
  title={Sinkhorn distances: Lightspeed computation of optimal transport},
  author={Cuturi, Marco},
  journal={Advances in neural information processing systems},
  volume={26},
  year={2013}
}

@inproceedings{zhang2025adapting,
  title={Adapting dense matching for homography estimation with grid-based acceleration},
  author={Zhang, Kaining and Deng, Yuxin and Ma, Jiayi and Favaro, Paolo},
  booktitle={Proceedings of the Computer Vision and Pattern Recognition Conference},
  pages={6294--6303},
  year={2025}
}

@inproceedings{brown2011multi,
  title={Multi-spectral SIFT for scene category recognition},
  author={Brown, Matthew and S{\"u}sstrunk, Sabine},
  booktitle={CVPR 2011},
  pages={177--184},
  year={2011},
  organization={IEEE}
}

@article{sun2022drone,
  title={Drone-based RGB-infrared cross-modality vehicle detection via uncertainty-aware learning},
  author={Sun, Yiming and Cao, Bing and Zhu, Pengfei and Hu, Qinghua},
  journal={IEEE Transactions on Circuits and Systems for Video Technology},
  volume={32},
  number={10},
  pages={6700--6713},
  year={2022},
  publisher={IEEE}
}

@inproceedings{detone2018superpoint,
  title={Superpoint: Self-supervised interest point detection and description},
  author={DeTone, Daniel and Malisiewicz, Tomasz and Rabinovich, Andrew},
  booktitle={Proceedings of the IEEE conference on computer vision and pattern recognition workshops},
  pages={224--236},
  year={2018}
}

@inproceedings{tian2019sosnet,
  title={Sosnet: Second order similarity regularization for local descriptor learning},
  author={Tian, Yurun and Yu, Xin and Fan, Bin and Wu, Fuchao and Heijnen, Huub and Balntas, Vassileios},
  booktitle={Proceedings of the IEEE/CVF conference on computer vision and pattern recognition},
  pages={11016--11025},
  year={2019}
}

@inproceedings{dusmanu2019d2,
  title={D2-net: A trainable cnn for joint description and detection of local features},
  author={Dusmanu, Mihai and Rocco, Ignacio and Pajdla, Tomas and Pollefeys, Marc and Sivic, Josef and Torii, Akihiko and Sattler, Torsten},
  booktitle={Proceedings of the ieee/cvf conference on computer vision and pattern recognition},
  pages={8092--8101},
  year={2019}
}

@inproceedings{wang2024efficient,
  title={Efficient LoFTR: Semi-dense local feature matching with sparse-like speed},
  author={Wang, Yifan and He, Xingyi and Peng, Sida and Tan, Dongli and Zhou, Xiaowei},
  booktitle={Proceedings of the IEEE/CVF conference on computer vision and pattern recognition},
  pages={21666--21675},
  year={2024}
}

@article{holland1977robust,
  title={Robust regression using iteratively reweighted least-squares},
  author={Holland, Paul W and Welsch, Roy E},
  journal={Communications in Statistics-theory and Methods},
  volume={6},
  number={9},
  pages={813--827},
  year={1977},
  publisher={Taylor \& Francis}
}

@inproceedings{barath2019magsac,
  title={MAGSAC: marginalizing sample consensus},
  author={Barath, Daniel and Matas, Jiri and Noskova, Jana},
  booktitle={Proceedings of the IEEE/CVF conference on computer vision and pattern recognition},
  pages={10197--10205},
  year={2019}
}

@inproceedings{edstedt2024roma,
  title={Roma: Robust dense feature matching},
  author={Edstedt, Johan and Sun, Qiyu and B{\"o}kman, Georg and Wadenb{\"a}ck, M{\aa}rten and Felsberg, Michael},
  booktitle={Proceedings of the IEEE/CVF Conference on Computer Vision and Pattern Recognition},
  pages={19790--19800},
  year={2024}
}

@article{liu2024codinghomo,
  title={Codinghomo: Bootstrapping deep homography with video coding},
  author={Liu, Yike and Li, Haipeng and Liu, Shuaicheng and Zeng, Bing},
  journal={IEEE Transactions on Circuits and Systems for Video Technology},
  volume={34},
  number={11},
  pages={11214--11228},
  year={2024},
  publisher={IEEE}
}

@article{yang2021siamcorners,
  title={SiamCorners: Siamese corner networks for visual tracking},
  author={Yang, Kai and He, Zhenyu and Pei, Wenjie and Zhou, Zikun and Li, Xin and Yuan, Di and Zhang, Haijun},
  journal={IEEE Transactions on Multimedia},
  volume={24},
  pages={1956--1967},
  year={2021},
  publisher={IEEE}
}

@inproceedings{xie2025p2wnet,
author = {Xie, ShangXuan and Wu, HaiFeng and Li, Wen and Duan, Lixin},
year = {2025},
month = {06},
pages = {1-6},
title = {P2WNet: Homography Estimation for Part-To-Whole and Cross-Modality Scenarios},
doi = {10.1109/ICME59968.2025.11209935}
}

@article{zhao2021image,
  title={Image stitching via deep homography estimation},
  author={Zhao, Qiang and Ma, Yike and Zhu, Chen and Yao, Chunfeng and Feng, Bailan and Dai, Feng},
  journal={Neurocomputing},
  volume={450},
  pages={219--229},
  year={2021},
  publisher={Elsevier}
}

@article{roy2014landsat,
  title={Landsat-8: Science and product vision for terrestrial global change research},
  author={Roy, David P and Wulder, Michael A and Loveland, Thomas R and Ce, Woodcock and Allen, Richard G and Anderson, Martha C and Helder, Dennis and Irons, James R and Johnson, David M and Kennedy, Robert and others},
  journal={Remote sensing of Environment},
  volume={145},
  pages={154--172},
  year={2014},
  publisher={Elsevier}
}

@article{xie2024rcvs,
  title={Rcvs: A unified registration and fusion framework for video streams},
  author={Xie, Housheng and Sang, Meng and Zhang, Yukuan and Yang, Yang and Zhao, Shan and Zhong, Jianbo},
  journal={IEEE Transactions on Multimedia},
  volume={26},
  pages={11031--11043},
  year={2024},
  publisher={IEEE}
}

@inproceedings{bradley2021cinematic,
  title={Cinematic-L1 video stabilization with a log-homography model},
  author={Bradley, Arwen and Klivington, Jason and Triscari, Joseph and van der Merwe, Rudolph},
  booktitle={Proceedings of the IEEE/CVF winter conference on applications of computer vision},
  pages={1041--1049},
  year={2021}
}

@article{xu2022dut,
  title={Dut: Learning video stabilization by simply watching unstable videos},
  author={Xu, Yufei and Zhang, Jing and Maybank, Stephen J and Tao, Dacheng},
  journal={IEEE Transactions on Image Processing},
  volume={31},
  pages={4306--4320},
  year={2022},
  publisher={IEEE}
}

@article{zhao2020fast,
  title={Fast georeferenced aerial image stitching with absolute rotation averaging and planar-restricted pose graph},
  author={Zhao, Yong and Liu, Guochen and Xu, Shibiao and Bu, Shuhui and Jiang, Hongkai and Wan, Gang},
  journal={IEEE Transactions on Geoscience and Remote Sensing},
  volume={59},
  number={4},
  pages={3502--3517},
  year={2020},
  publisher={IEEE}
}

@article{li2024seam,
  title={Seam-Adaptive Structure-Preserving Image Stitching for Drone Images},
  author={Li, Jiaxue and Zhou, Yicong},
  journal={IEEE Transactions on Geoscience and Remote Sensing},
  volume={63},
  pages={1--12},
  year={2024},
  publisher={IEEE}
}

@article{zarei2022megastitch,
  title={MegaStitch: Robust Large-scale image stitching},
  author={Zarei, Ariyan and Gonzalez, Emmanuel and Merchant, Nirav and Pauli, Duke and Lyons, Eric and Barnard, Kobus},
  journal={IEEE Transactions on Geoscience and Remote Sensing},
  volume={60},
  pages={1--9},
  year={2022},
  publisher={IEEE}
}

@article{li2023multimodal,
  title={Multimodal image fusion framework for end-to-end remote sensing image registration},
  author={Li, Liangzhi and Han, Ling and Ding, Mingtao and Cao, Hongye},
  journal={IEEE Transactions on Geoscience and Remote Sensing},
  volume={61},
  pages={1--14},
  year={2023},
  publisher={IEEE}
}

@article{zhou2024uncertainty,
  title={Uncertainty guided deep Lucas-Kanade homography for multimodal image alignment},
  author={Zhou, Zhen and Luo, Jianqiao and Zhu, Qing and Wang, Yaonan and Zhong, Hang and Feng, Mingtao and Chen, Lin},
  journal={IEEE Transactions on Geoscience and Remote Sensing},
  volume={63},
  pages={1--14},
  year={2024},
  publisher={IEEE}
}

@article{wang2025integrated,
  title={An Integrated Inter-Frame Stabilization and Fast Imaging Method for Video Synthetic Aperture Radar},
  author={Wang, Shuo and Wang, Guanyong and Wang, Yanping and Zhou, Rui and Zhao, Meng and Wang, Yuxiao},
  journal={IEEE Transactions on Geoscience and Remote Sensing},
  year={2025},
  publisher={IEEE}
}

@article{linger2014aerial,
  title={Aerial image registration for tracking},
  author={Linger, Michael E and Goshtasby, A Ardeshir},
  journal={IEEE Transactions on Geoscience and Remote Sensing},
  volume={53},
  number={4},
  pages={2137--2145},
  year={2014},
  publisher={IEEE}
}

@article{xue2023smalltrack,
  title={Smalltrack: Wavelet pooling and graph enhanced classification for uav small object tracking},
  author={Xue, Yuanliang and Jin, Guodong and Shen, Tao and Tan, Lining and Wang, Nian and Gao, Jing and Wang, Lianfeng},
  journal={IEEE Transactions on Geoscience and Remote Sensing},
  volume={61},
  pages={1--15},
  year={2023},
  publisher={IEEE}
}

\end{document}